\documentclass{article}

\usepackage[preprint]{arxiv}

\usepackage[utf8]{inputenc} 
\usepackage[T1]{fontenc}    
\usepackage{hyperref}       
\usepackage{url}            
\usepackage{booktabs}       
\usepackage{amsfonts}       
\usepackage{nicefrac}       
\usepackage{microtype}      

\usepackage{url}            
\usepackage{booktabs}       
\usepackage{amsfonts}       
\usepackage{amsmath}
\usepackage{amssymb}
\usepackage{nicefrac}       
\usepackage{microtype}      
\usepackage{csquotes}

\usepackage{paralist, amsmath, amssymb, graphicx}
\usepackage{amssymb}
\usepackage{mathtools}
\usepackage{nccmath}
\usepackage[table,xcdraw,dvipsnames]{xcolor}
\usepackage{tabularray}

\usepackage{nicefrac}
\usepackage{stmaryrd}
\usepackage{bm}
\usepackage{cancel}
\usepackage{multirow}
\usepackage{tablefootnote}

\usepackage{xfrac}
\usepackage{mathrsfs}
\usepackage{amscd}
\usepackage{dsfont}
\usepackage{tikz}
\usepackage{epstopdf}
\usepackage{enumerate}
\usepackage{bm}
\usepackage{chngcntr} 
\usepackage{makecell}
\usepackage{placeins}
\usepackage{algorithm}
\usepackage{algorithmic}
\usepackage{multicol}
\usepackage{xargs}
\usepackage{soul}

\usepackage{adjustbox}
\usepackage{tocloft}
\usepackage{pifont}
\usepackage[labelfont=bf]{caption}
\usepackage{natbib}
\usepackage{enumitem}

\usepackage{todonotes}

\definecolor{LightGray}{gray}{0.8}
\definecolor{LightBlue}{rgb}{0.0, 0.7, 1.0}

\newtheorem{definition}{Definition}

\title{
Forging Time Series with Language: A Large Language Model Approach to Synthetic Data Generation}

\author{%
  Cécile Rousseau  \phantom{ii}\\
  IBM Research Europe  \phantom{ii}\\
 \texttt{rousseau.cecile@ibm.com} \phantom{ii}\\
  \And
   Tobia Boschi\\
  IBM Research Europe \\
  \phantom{iii} \texttt{tobia.boschi@ibm.com} \phantom{iii}\\
  \And
  Giandomenico Cornacchia \\
  IBM Research Europe\\
  \texttt{Giandomenico.Cornacchia1@ibm.com} \\
  \And
  Dhaval Salwala \\
  IBM Research Europe\\
  \texttt{dhaval.vinodbhai.salwala@ibm.com} \\
  \AND
  Alessandra Pascale \\
  IBM Research Europe\\
  \phantom{iiiiiii} \texttt{apascale@ie.ibm.com} \phantom{iiiiiii} \\
  \And
  Juan Bernabe Moreno  \\
  IBM Research Europe\\
  \texttt{juan.bernabe-moreno@ibm.com} \\
}

\begin{document}
\maketitle


\begin{abstract}
SDForger is a flexible and efficient framework for generating high-quality multivariate time series using LLMs.
Leveraging a compact data representation, SDForger provides synthetic time series generation from a few samples and low-computation fine-tuning of any autoregressive LLM. Specifically, the framework transforms univariate and multivariate signals into tabular embeddings, which are then encoded into text and used to fine-tune the LLM.
At inference, new textual embeddings are sampled and decoded into synthetic time series that retain the original data's statistical properties and temporal dynamics. Across a diverse range of datasets, SDForger outperforms existing generative models in many scenarios, both in similarity-based evaluations and downstream forecasting tasks. By enabling textual conditioning in the generation process, SDForger paves the way for multimodal modeling and the streamlined integration of time series with textual information. The model is open-sourced at \url{https://github.com/IBM/fms-dgt/tree/main/fms_dgt/public/databuilders/time_series}.
\end{abstract}

%
%

\section{Introduction}
\label{sec:intro}

In the era of foundation models, integrating time-series analysis with language models (LLMs) has emerged as a key research priority, spurring work on specialized models for forecasting, anomaly detection and root cause analysis \citep{liang2024foundation}. These efforts aim to harness the representational power of LLMs to tackle complex temporal dependencies. Despite these advancements, foundation models for time series still struggle with concept drift, high variability, and face performance degradation over long horizons. In addition, the scarcity of large and diverse time-series datasets often limits model generalization, especially in domains where data collection is expensive or operationally constrained, such as climate science, finance, or plasma physics~\citep{kit2024learning}.

In this context, \textit{synthetic data generation has emerged as a complementary research direction to improve scalability and model performance, particularly in low-quality or data-scarce settings.}
In particular, synthetic data can be leveraged to fine-tune machine learning models
on domain-specific distributions, reducing the need to train models from scratch and improving sample efficiency. Indeed, several generative models for time series have been proposed, including VAE-based architectures~ \citep{desai2021timevae, lee2023vector}, GANs~\citep{pei2021towards, kidger2021neural}, and diffusion-based approaches~\citep{zhou2023deep}. While these methods have shown promise, they typically lack pretraining, require task-specific retraining, and often struggle with long-term dependencies, multivariate coupling, and distributional shifts~\citep{ang2023tsgbench}.

Recent studies have instead explored the use of LLMs for synthetic tabular data generation~\citep{padhi2021tabular, borisov2022language}, demonstrating that language models trained on text-encoded data can capture statistical relationships and feature dependencies effectively. However, extending these methods to time series is non-trivial: long temporal windows increase inference and training costs, while temporal and multivariate correlations require more structured modeling.

These limitations highlight the need for novel methodologies that adapt LLMs for time-series generation while addressing temporal, structural, and computational challenges. We introduce \textbf{\textit{SDForger}} (Synthetic Data Forger), a novel framework for generating high-quality \emph{univariate and multivariate time series}, even in data-scarce settings. SDForger leverages foundation models with minimal fine-tuning by operating over compact tabular embeddings derived from functional decompositions.

\textbf{Key features and advantages of SDForger:}
\begin{itemize}[leftmargin=0.6cm, noitemsep, topsep=0pt]

\item \textbf{Compact basis representation} SDForger uses FastICA or PCA to embed time series into low-dimensional tabular data. These capture key temporal and inter-variable structures while decoupling the embedding from sequence length, enabling efficient processing of long signals.

\item \textbf{Text-to-sequence generation via LLMs} The embedding tables are converted into structured textual prompts and used to fine-tune a language model. A guided inference approach is then used to generate structured embeddings, ensuring that the synthetic data retains the original dataset's statistical properties and feature relationships.


\item \textbf{Flexible, lightweight architecture} The framework leverages autoregressive LLMs, including lightweight models, and requires only a small number of training instances.
Its modular design enables easy adaptation to different generation tasks and architectures, while its compact embedding space ensures fast inference, even for long time-series windows.

\item \textbf{Multivariate and multimodal readiness} SDForger can model complex multivariate dynamics and supports future extensions to textual conditioning, enabling generation guided by both time-series structure and external language-based context.

\end{itemize}

By combining structured embeddings with LLM-based generation, SDForger establishes a new paradigm for scalable, interpretable, and high-quality synthetic time-series generation.

Our simulations demonstrate that SDForger not only generates statistically realistic time series but also improves downstream model performance, often matching or exceeding results obtained from real data alone. This is particularly valuable in practical scenarios where access to high-quality data is limited or where distribution shifts make original training data less effective. Compared to state-of-the-art generative models, SDForger achieves competitive or superior performance across a wide range of similarity metrics and utility-based evaluations.
Notably, our experiments show that even lightweight, pretrained LLMs (e.g., GPT-2) are sufficient to produce high-quality synthetic data with minimal fine-tuning, highlighting the accessibility, efficiency, and flexibility of our approach.

In the remainder of this paper, we first review related work (Section~\ref{sec:RelatedWork}), then detail the SDForger framework (Section~\ref{sec:Methodology}), describe our evaluation setup (Section~\ref{sec:EvaluationSettings}), present extensive experimental results (Section~\ref{sec:Results}), highlight the flexibility of language models (Section~\ref{sec:TextConditional}), and conclude with key takeaways and future directions (Section~\ref{sec:Conclusions}).

\section{Related work}
\label{sec:RelatedWork}

\paragraph{Time-series generation}
Recent advances in time-series generation have introduced a variety of deep generative models, including GAN-based approaches like TimeGAN~\citep{smith2020conditional}, state-space models, and vector-quantized architectures such as TimeVQVAE~\citep{lee2023vector}. While these methods generate realistic sequences, they often struggle with multivariate dependencies, and require training from scratch for each dataset. More recent framework integrate randomly-weighted combinations of time series to improve their pretraining pipeline, e.g. Chronos~\citep{ansari2024chronos} adapting the Mixup~\citep{zhou2023improving} methodology for time series.
To address evaluation challenges, TSGBench~\citep{ang2023tsgbench} proposes a unified benchmark of similarity, fidelity, and utility metrics.

\paragraph{Foundation models and LLMs for time series}
Recent advances have introduced foundation models specifically designed for time-series tasks, offering unified frameworks for forecasting, classification, and anomaly detection~\citep{liang2024foundation}. Examples include TFT~\citep{lim2021temporal}, TimeGPT~\citep{garza2023timegpt}, and Chronos~\citep{ansari2024chronos}, while lightweight models like TTMs~\citep{ekambaram2024ttms} focus on efficient multivariate forecasting. Parallel efforts have explored adapting LLMs to time-series data by encoding numerical sequences as text~\citep{gruver2024large, zhou2023one, jin2023time}, enabling zero-shot inference and transfer learning. Notably, LLMTime~\citep{gruver2024large} and GPT4TS~\citep{zhou2023one} retain most of the LLM architecture while fine-tuning only shallow layers, and Time-LLM~\citep{jin2023time} employs reprogramming to adapt to temporal tasks. Despite these innovations, most existing approaches either require task-specific pretraining or struggle to model complex 
structures and may benefit from synthetic data for improving generalization and robustness.

\paragraph{Specialized architectures for time-series generation}
Several architectures have been specifically designed to capture temporal and multivariate dependencies in time series. Variational autoencoders such as TimeVAE~\citep{desai2021timevae} and TimeVQVAE~\citep{lee2023vector} use recurrent or vector-quantized structures to model sequential dynamics. GAN-based approaches, including RTSGAN~\citep{pei2021towards}, SDEGAN~\citep{kidger2021neural}, and COSCI-GAN~\citep{seyfi2022generating}, employ recurrent or component-wise disentangled generators to capture complex temporal patterns adversarially. Diffusion-based models, like LS4~\citep{zhou2023deep}, generate sequences through learned reverse-time processes. These specialized architectures complement general-purpose time-series generation methods and provide valuable baselines for evaluating synthetic data. However, these architectures are trained from scratch and cannot leverage existing pre-trained language or foundation models, limiting their scalability and adaptability across domains.

%
%

\section{Methodology}
\label{sec:Methodology}

\begin{figure}[ht!]
  \centering
  \includegraphics[width=0.95\linewidth]{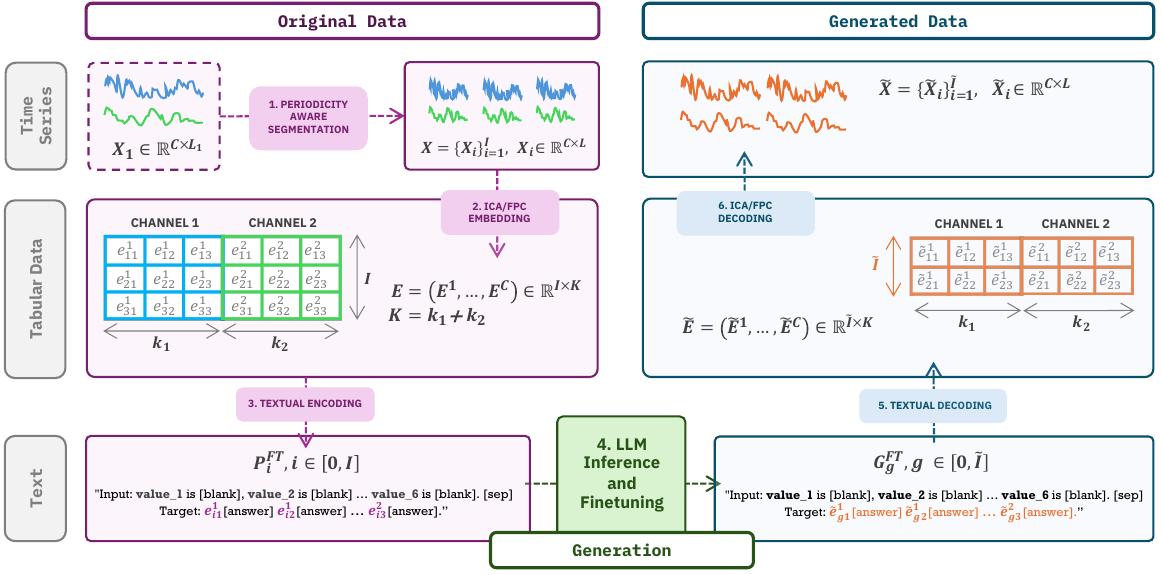} 
 \caption{\textbf{SDForger pipeline.} Overview of the SDForger generation process. The example illustrates a setting with $I=3$ input segments, $C=2$ channels, $k_1 = k_2 = 3$ components, and $\tilde{I} = 2$ generated samples. The model performs periodicity-aware segmentation, extracts embeddings, and embed them into text. An LLM is then fine-tuned to generate embedding sequences, which are finally decoded to reconstruct synthetic time series.}
  \label{fig:pipeline}
\end{figure}

In this section, we present our methodology, illustrated in Figure~\ref{fig:pipeline}. SDForger is divided into three macro-steps: (i) \emph{Preprocessing and embedding} transform the time series into tabular data (i.e., steps 1 to 3 highlighted in purple); (ii) \textit{Fine-tuning and Generation} fine-tune a pre-trained LLM and generate new embedding instances (i.e., step 4 highlighted in green); (iii) \textit{Decoding} reconstruct the original time-series space from the generated embeddings (i.e., steps 5 and 6 highlighted in light blue).  

\paragraph{Notation}
Hereinafter, we introduce some basic notation. Let \( X = \{X_i\}_{i=1}^I, \ {X_i} \in \mathbb{R}^{C \times L} \) represent a collection of \( I \) instances of a multivariate time series, where each instance has length \( L \) and consists of \( C \) channels. The task of \textit{synthetic time-series generation} can now be formally defined as producing \( \tilde{I} \) instances of a multivariate time series \( \{\tilde{X}_i\}_{i=1}^{\tilde{I}}, \tilde{X}_i \in \mathbb{R}^{C \times L} \) conditioned on the given context \( X \). Throughout the paper, we denote by \( x_i^c \in \mathbb{R}^L \) the \( i \)-th instance of channel \( c \), and by \( X^c \in \mathbb{R}^{I \times L} \) the matrix collecting all instances associated with channel \( c \).

\paragraph{Periodicity-aware segmentation}
In cases where only a single time series instance \( X_1 \in \mathbb{R}^{C \times L_1} \) is available, we apply a segmentation strategy to artificially create multiple instances. The segmentation procedure facilitates the estimation of the time-series distribution. Specifically, we extract \( I \) periodicity-aware windows of fixed length \( L < L_1 \), aligning cuts with natural cycles and minimizing overlap to enhance independence and diversity. This pre-processing (described in Appendix~\ref{appendix:data_preprocessing}) transforms the data from a single sequence \( X_1 \) into a set \( X = \{X_i\}_{i=1}^I \), where \( X_i \in \mathbb{R}^{C \times L} \), preparing the data for the generation task.

\subsection{From time series to tabular data}
\label{subsec:embeddings}


To enable tabular generation and analysis, SDForger transforms time series into structured tabular data using \textit{basis decomposition techniques}. Each row represents a time series embedding obtained by projecting the signal onto a set of learned basis functions.
Specifically, we adopt two decomposition methods: Functional Principal Components (FPC) \citep{ramsay2005} and Fast Independent Component Analysis (FastICA) \citep{hyvarinen1999fast}:


\begin{itemize}[leftmargin=0.6cm, noitemsep, topsep=0pt]
    \item \textbf{FPC} identifies principal modes of variation by performing eigen-decomposition of the covariance operator. It captures directions of maximal variance, preserving correlation across components, showing effectiveness in modeling multivariate longitudinal data~\citep{boschi2024fasten}.
    \item \textbf{FastICA} extracts statistically independent components by maximizing non-Gaussianity. It decomposes a contrast function, uncovering independent latent factors that may not align with the directions of maximal variance.
\end{itemize}
Formally, for each channel \( c \), we assume the instances \( (X^c_1, \dots, X^c_I) \) are realizations of continuous functions defined over \( \mathcal{T} = [0, L] \). 
We approximate each \( X^c_i \) as a linear combination of \( k_c \) basis functions \( (b^c_1, \dots, b^c_{k_c}) \), where the choice of basis depends on the decomposition method (non-Gaussianity-based for FastICA, covariance-based for FPC). 
The embedding coefficients are:
\begin{align*}
    e^c_{ij} = \langle X^c_i, b^c_j \rangle_{\mathbb{L}^2} = \int_\mathcal{T} X^c_i(t) b^c_j(t) \, \mathrm{d}t,
\end{align*}
We define the embedding matrix for channel \( c \), \( E^c \in \mathbb{R}^{I \times k_c} \). 
By concatenating the embeddings across all channels, we obtain the final embedding table:
$ E = (E^1, \dots, E^C) \in \mathbb{R}^{I \times K}$ with $K = \sum_{c=1}^C k_c.$

Throughout this paper, we refer to the columns of \( E \) as embedding features. 
We denote the \( i \)-th row of \( E \) as \( E_i \), corresponding to the embedding vector of instance \( X_i \), and the value in its \( k \)-th column as \( e_{ik} \). 
More details on the choice of \( k_c \) are given in Appendix~\ref{appendix:k_selection} and Appendix Table~\ref{tab:variance_fica_fpc}.

Notably, both methods offer the advantage that their computational cost depends on the number of instances \( I \) and the number of components \( k_c \), but not on the instance length \( L \). 
This decoupling allows our algorithm to handle very long time windows without a corresponding increase in computational complexity, ensuring great flexibility and scalability.


\subsection{Generation of tabular data} 

Our data generation block consists of three key stages: encoding tabular data into text, fine-tuning an LLM, and generating synthetic embeddings. 

\subsubsection{From embeddings table to text} 
LLMs are designed to process textual information. Therefore, applying an LLM to tabular data requires converting each row into a textual format that can serve as a prompt during the fine-tuning stage. Inspired by \citet{donahue2020enabling}, we introduce a \textit{Textual Encoder} responsible for converting tabular instances $E_i$ into structured text representations using a Fill-In-The-Middle template.


\begin{definition}[Textual encoder]  
Let $\mathcal{P}^{\textup{FT}}=\{{\mathcal{P}_i^{\textup{FT}}\}}_{i=1}^{I}$ denote the set of fine-tuning prompts, where:

\begin{align*}
\mathcal{P}_i^{\textup{FT}} = \texttt{``} & \texttt{Input:\ }
\bigcirc_{k=1}^K {\texttt{(value\_{$\pi(k)$} is [blank],)}} \texttt{\ [sep]\ } \\
& \texttt{Target:\ } \bigcirc_{k=1}^K { \texttt{($e_{i\pi(k)}$ [answer])''} }
\end{align*}
Here, the operator $ \bigcirc $ denotes the concatenation and $ \pi $
is a random permutation of \( K \) elements.
\end{definition} 

\emph{Random Feature Order Permutation}.  
Encoding tabular data into text can introduce unintended positional biases, as LLMs inherently process tokens in sequence. To enforce order independence \citep{borisov2022language}, we apply a random permutation $\pi$ to the encoded feature-value pairs within each instance. This shuffling ensures that the model does not infer any spurious relationships based on the ordering of features within the textual representation. For $K=2$, an admissible finetuning prompt for $E_i$ is:
\textit{\texttt{``Input:\ value\_2 is [blank], value\_1 is [blank]\ [sep]\ Target:\ $e_{i2}$ [answer] $e_{i1}$ [answer]''}}


\subsubsection{Large language model finetuning and inference} 


\paragraph{Fine-tuning} 
By training an LLM on structured text representations of the embedding tables, we enable it to learn meaningful patterns present in the data. Since the optimal number of fine-tuning epochs depends on the number of instances, the embedding dimension, and the LLM architecture, we implement an \textit{early stopping criterion} to prevent overfitting. 

\paragraph{Inference}
After fine-tuning, inference is performed by prompting the LLM with structured textual templates that mirror the training format, allowing it to autonomously generate new embedding rows. 

\begin{definition}[Textual inference]
Given the embedding table $ E \in \mathbb{R}^{I \times K} $, we define the set of inference prompts at each generation step as $\mathcal{P}^{\textup{INF}} = \{ {\mathcal{P}_g^{\textup{INF}}\}}_{g=1}^{G}$ where:
\begin{align*}
\mathcal{P}_g^{\textup{INF}} = \texttt{``} & \texttt{Input:}
\bigcirc_{k=1}^K {\texttt{( value\_{$\pi(k)$} is [blank], ) } } \texttt{ [sep] } \texttt{Target:'' }
\end{align*}
\end{definition}

We use a multinomial distribution sampling strategy to reduce repetition and generate more creative and diverse outputs. The model draws from its learned token probability distribution at each step, guided by the temperature parameter, which controls sampling variability. As a result, all values are internally generated by the LLM in a fully conditional and self-contained manner, highlighting the model capacity to internalize statistical and structural patterns from compact embeddings and synthesize coherent time series without external noise injection or sampling routines.

At each inference step, we generate a batch of $G$ synthetic instances, repeating the process until the desired number of sequences is obtained or a stopping criterion is met. We denote the set of all generated text instances as:
$ \mathcal{G} = \{\mathcal{G}_1, \dots, \mathcal{G}_G \}. $
Ideally, the fine-tuned LLM should generate text instances in the following format: $\mathcal{G}_g = \bigcirc \  ( \ \mathcal{P}_g^{\textup{INF}} , \ \bigcirc_{k=1}^K ( \text{$\tilde{a}_{g \pi(k)}$ \textit{\texttt{[answer]}}} \ )  \ )$ where $ \pi $ is the random permutation used in $\mathcal{P}_g^{\textup{INF}}$, and $ \{ \tilde{a}_{g\pi(k)} \}_{k=1}^{K} $ are the $ K $ numerical values inferred, which form the generated embedding table.

\paragraph{Retrieve embedding from text}

Given a generated text instance $\mathcal{G}_g \in \mathcal{G}$, we reconstruct the corresponding tabular data by mapping the inferred embedding values $ \{ \tilde{e}_{g\pi(k)} \}_{k=1}^{K} $ to their respective features. Each textual entry is split into feature-value pairs using \textit{\texttt{"[answer]"}} as a delimiter. Missing or unrecognized features are assigned the placeholder \texttt{"NaN"}. For a specific channel $ c $, the output of an inference step $ s $ is the reconstructed embedding matrix:
$
\tilde{E}^{c,s} \in \mathbb{R}^{G \times k_c},
$
where each row corresponds to a generated instance $ \mathcal{G}_g $ and each column represents an inferred embedding feature associated with channel $ c $. To track all generated embeddings up to step $s$, we define: $\tilde{E}^{c,\le s}$.

\paragraph{In-generation filtering and stopping criterion}

At each inference step \( s \), we apply an online filtering procedure that validates generated embeddings without requiring reconstruction into the time-series domain, ensuring efficient real-time evaluation. Specifically, the reconstructed embedding matrices \( \big( \tilde{E}^{1, s}, \dots, \tilde{E}^{C, s} \big) \) are filtered based on three criteria: 1) Instances with missing values are discarded, as they prevent accurate reconstruction; 2) Duplicated instances are discarded to maintain diversity in the generated dataset; 3) Significantly diverging instances are discarded. This combined filtering procedure not only enforces diversity and validity among the generated instances but also provides a diagnostic signal: if a substantial fraction of samples is rejected, it may indicate that the fine-tuned LLM requires further training or more representative data.
Representative examples of discarded instances and details on the divergence detection procedure are provided in Appendix~\ref{appendix:filtering}, while Appendix~Table~\ref{tab:filtering_rejection} reports the rejection rates observed in a representative generation scenario, illustrating the balance between filtering rigor and sample diversity.

In generation mode, SDForger employs a dynamic stopping criterion that continues generating batches of \( G \) text instances as long as sufficient diversity is preserved among the generated samples (Appendix~\ref{appendix:stopping}). However, for consistent comparison with baseline methods across all simulation scenarios, we fix the number of generated instances \( \tilde{I} \) across all algorithms. If we denote by \( S \) the final inference step, then the output of the generation process is the complete embedding table \( \tilde{E} \in \mathbb{R}^{\tilde{I} \times K} \), where, for each channel \( c \), \( \tilde{E}^c = \tilde{E}^{c, \leq S} \).

\subsection{Decoding: from tabular embeddings to time series}

Given $ \tilde{E}$, the time-series representation of generated embeddings can be efficiently recovered due to the reversible nature of the embedding technique used. 
For the channel $c$, given the generated coefficients $\tilde e^c_{ij}$ and the corresponding basis system $(b^c_1, \dots, b^c_{k_c})$, the reconstructed time series are computed as follows \cite{kokoszka2017introduction}: $ \tilde{x}^c_i = \sum_{j=1}^{k_{c}} \tilde e^c_{ij} b^c_j $.


This formulation ensures that each generated embedding is decoded back to the original space, resulting in $\tilde I$ synthetic instances of a multivariate time series $\tilde{X}_i \in \mathbb{R}^{C \times L}$.

%
%
\section{Evaluation methodology}
\label{sec:EvaluationSettings}


\paragraph{Evaluation metrics} Evaluating synthetic time-series data requires balancing \textit{realism}, \textit{usability}, and \textit{efficiency}. A strong generative model should replicate key properties of real data while supporting downstream tasks such as forecasting. We adopt a comprehensive evaluation framework comprising two categories: \textit{similarity metrics} and \textit{utility metrics}.

\begin{itemize}[leftmargin=0.6cm, noitemsep, topsep=0pt]
\item \textbf{Similarity metrics}, inspired by \citet{ang2023tsgbench}, assess how closely the generated data matches the real data in terms of distribution, structure, and behavior. They fall into two subtypes:
(i) \textbf{Feature-based metrics} which include \textit{Marginal Distribution Difference (MDD)}, \textit{Auto-Correlation Difference (ACD)}, \textit{Skewness Difference (SD)}, and \textit{Kurtosis Difference (KD)}, assess how well synthetic data retains key statistical properties of real data;
 (ii) \textbf{Distance-based metrics} include \textit{Euclidean Distance (ED)}, \textit{Dynamic Time Warping (DTW)}, and \textit{SHAP-RE} (SHR), a shapelet-based reconstruction error. They quantify the similarity between synthetic and real data in raw feature space or temporal alignment. Formal definitions are provided in Appendix~\ref{appendix:evaluation_metrics}.
 \item {Utility metrics} assess the effectiveness of synthetic data in downstream tasks. Specifically, we fine-tune Tiny Time Mixers (TTM) \citep{ekambaram2024ttms}, a recent foundation model for multivariate time series, under four settings: (1) zero-shot (no fine-tuning) (2) real data only, (3) synthetic data only, and (4) real data augmented with synthetic data. This setup quantifies the impact of synthetic data on model transferability, data efficiency, and robustness. 
\end{itemize}

\paragraph{Evaluation protocols} 

We consider three distinct evaluation settings to assess the generative capabilities of SDForger across different structural assumptions:
\begin{itemize}[leftmargin=0.6cm, noitemsep, topsep=0pt]
    \item \textbf{Multisample generation} aims to produce new instances by combining patterns from multiple existing time series. This setting reflects scenarios such as generating experimental samples, weather profiles, or patient trajectories from heterogeneous observations. It emphasizes diversity and generalization in data-rich contexts.
    \item \textbf{Univariate generation} focuses on learning from a single time series to generate plausible alternative versions. This is useful for simulating counterfactual histories, seasonal variations, or stress-test scenarios in domains like finance, weather, and demand forecasting.
    \item \textbf{Multivariate generation} evaluates the ability to jointly generate multiple interdependent channels. It reflects real-world settings, such as energy systems, traffic flows, or sensor networks, where channel interactions and cross-correlations are crucial for realism and downstream utility.
\end{itemize}
In the multisample case, multiple instances are available by design. In contrast, for univariate and multivariate settings, only one instance is provided; therefore, we first apply the period-aware segmentation procedure described in Section~\ref{sec:Methodology} to extract multiple windows from each channel.

\paragraph{Parameter settings} 

We summarize here the hyperparameters for SDForger. We fix the embedding dimension to $\bm{k=3}$ for the \textit{multisample} and \textit{univariate setting}. The LLM used for generation is GPT-2 \footnote{\url{https://huggingface.co/openai-community/gpt2}}, fine-tuned with Adam~\citep{diederik2014adam} optimization, a learning rate of $8 \times 10^{-5}$, batch size 32, and a maximum of 200 epochs. Early stopping criteria is applied based on the best validation loss computed every 5 steps, patience set to 5, randomly choosing $20\%$ of the data as a validation set. 

    
    

\paragraph{Baselines} We evaluated SDForger's performance against several baseline models for synthetic time series generation, covering different approaches. \emph{Variational autoencoders}: TimeVAE \citep{desai2021timevae}, which models temporal dependencies with a recurrent VAE architecture, and TimeVQVAE \citep{lee2023vector}, which incorporates vector quantization for better capturing discrete temporal patterns; \emph{generative adversarial networks}: RTSGAN \citep{pei2021towards}, which uses recurrent components for adversarial training, and SDEGAN \citep{kidger2021neural}, which models time series as solutions to stochastic differential equations; and a \emph{diffusion-based model}: LS4 \citep{zhou2023deep}, which generates sequences via a learned reverse-time diffusion process. Hyperparameters for all baseline competitors follow those reported in their original papers, except for SdeGAN, for which we fix the number of training iterations to $1000$ to balance convergence and computational cost.


\paragraph{Datasets} We evaluated SDForger models using 12 publicly available datasets from various domains, including energy, transport, industry, weather, and finance, with sampling frequencies ranging from 2 minutes to monthly. The datasets, sourced from the Monash Time Series Forecasting Repository and other public domains, include both stationary and non-stationary time series, reflecting diverse temporal dynamics. Detailed information is provided in Appendix~\ref{appendix:datasets}.

\section{Results}
\label{sec:Results}

Following, we discuss results on \textbf{Similarity-based} (Section~\ref{sec:sim_met_res}), \textbf{Utility-based metrics} (Section~\ref{sec:ut_met_res}), and a condensed ablation study (Section~\ref{sec:abl_c}). Complete ablations are provided in Appendix~\ref{appendix:additional_results}.

\subsection{Similarity-based metrics results}\label{sec:sim_met_res}

The similarity-based results aggregated for the multisample and univariate settings are reported in Table~\ref{tab:multisample_univariate_main_aggr}, with detailed per-dataset scores provided in Appendix Tables~\ref{tab:multisample_per_data}, \ref{tab:univariate_per_data_energy}, \ref{tab:univariate_per_data_transport}, \ref{tab:univariate_per_data_nature}, and \ref{tab:univariate_per_data_finance}. 


\textbf{Overall performance.} 
Different generative models exhibit complementary strengths: for instance, TimeVAE performs well on distribution-based metrics, while TimeVQVAE excels on distance-based measures such as Euclidean Distance and DTW. 
In contrast, \textit{SDForger achieves consistently strong and balanced performance across both metric categories}, maintaining high scores without overfitting to either statistical or structural similarity (Table~\ref{tab:multisample_univariate_main_aggr}). 
This balanced behaviour is further confirmed by the normalized average scores per metric group and the average rank values. 
Such consistency indicates that SDForger not only preserves key statistical features but also captures the underlying temporal and distributional structure of the data, demonstrating strong generalization and robustness across heterogeneous temporal domains. 
By decoupling representation learning from generation, SDForger captures long-range dependencies while maintaining statistical realism, ultimately producing temporally coherent and domain-consistent synthetic samples.


\textbf{Robustness to evaluation protocols} 
Comparing multisample and univariate settings, we observe that model rankings and relative performances remain largely consistent, suggesting that SDForger is robust to variations in the evaluation protocol. This stability is an important advantage in practice, where test-time conditions may vary.

\textbf{ICA vs. FPC} 
The ICA embedding strategy consistently leads, particularly on distance-based metrics. The superior performance of the ICA-based variant likely comes from the nature of the components it produces. Unlike FPC, which orders components by explained variance and often concentrates most information in the first few components, ICA explicitly seeks statistically independent components. This tends to produce a more balanced and disentangled basis decomposition, where each component carries distinct information that have similar importance for data reconstruction. For our LLM-based generation pipeline, this disentanglement appears advantageous because the model can learn a joint distribution over a set of factors that all have the same “power”. Thus, the results suggest that the LLM is indeed better equipped to model the joint distribution when presented with independent factors rather than a hierarchy of variance-ordered components. 

\begin{table}[t!]
\caption{\textbf{Aggregated performance comparison in the \textbf{multisample} and \textbf{univariate} settings.} Metrics include raw similarity scores and normalized averages (in $[0-1]$) for each metric group, plus the average rank. Lower values are better. \textbf{Bold} indicates the best performance per column, and \underline{underlined} indicates the second-best.}
\vspace{0.1cm}
\centering
\tiny
\renewcommand{\arraystretch}{1.4}
\rowcolors{2}{gray!20}{white}
\begin{tabular}{>{\columncolor{white}[-20pt][\tabcolsep]}l lrrrrrrrrrr}
    \cmidrule[1pt]{2-12}
    & & \multicolumn{4}{c}{\textbf{Feature-based}} & \multicolumn{3}{c}{\textbf{Distance-based}} & \multicolumn{2}{c}{\textbf{Norm. Avg.}} &  \\
    \cmidrule(lr){3-6} \cmidrule(lr){7-9} \cmidrule(lr){10-11} 
    & & \textbf{MDD} & \textbf{ACD} & \textbf{SD} & \textbf{KD} & \textbf{ED} & \textbf{DTW} & \textbf{SHR} & \textbf{Feat.} & \textbf{Dist.} & \textbf{Rank} \\
    \cmidrule[1pt]{2-12}
\multirow{7}{*}{\rotatebox[origin=c]{90}{\textbf{MULTISAMPLE}}}
& SDF-ICA$_3$ & 0.244 & \underline{1.180} & 0.869 & 2.384 & 16.669 & 12.373 & 6.870 & \underline{0.224} & 0.074 & 3.143 \\
& SDF-FPC$_{3}$ & 0.255 & 2.166 & 1.323 & 4.299 & 17.749 & 11.921 & 16.537 & 0.562 & 0.100 & 4.714 \\
& TimeVAE & \textbf{0.227} & \textbf{0.259} & \textbf{0.507} & \textbf{1.697} & 18.041 & \underline{11.625} & 14.021 & \textbf{0.000} & 0.094 & \textbf{2.143} \\
& TimeVQVAE & 0.371 & 5.466 & 1.327 & 3.889 & \textbf{13.661} & \textbf{10.167} & \textbf{2.030} & 0.873 & \textbf{0.000} & 3.714 \\
& RtsGAN & 0.279 & 1.769 & \underline{0.612} & \underline{2.300} & \underline{16.084} & 11.859 & \underline{5.631} & 0.231 & \underline{0.058} & \underline{2.857} \\
& SdeGAN & \underline{0.240} & 2.098 & 1.404 & 4.091 & 37.174 & 33.391 & 51.678 & 0.540 & 0.693 & 5.286 \\
& LS4 & 0.276 & 6.150 & 1.243 & 4.852 & 44.389 & 31.806 & 160.403 & 0.789 & 0.977 & 6.143 \\
\cmidrule(lr){2-12}
\multirow{7}{*}{\rotatebox[origin=c]{90}{\textbf{UNIVARIATE}}}
& SDF-ICA$_3$ & 0.306 & \textbf{1.396} & \underline{0.671} & 1.382 & \underline{18.802} & 12.435 & \underline{4.856} & \underline{0.149} & \underline{0.070} & \textbf{2.429} \\
& SDF-FPC$_{3}$ & 0.308 & \underline{1.480} & 0.801 & 1.690 & 19.340 & 12.809 & 5.452 & 0.354 & 0.084 & 4.000 \\
& TimeVAE & \underline{0.288} & 2.013 & \textbf{0.611} & \textbf{1.245} & 20.778 & \underline{12.126} & 18.534 & \textbf{0.066} & 0.158 & \underline{2.714} \\
& TimeVQVAE & 0.433 & 4.330 & 0.740 & 2.052 & \textbf{15.438} & \textbf{11.250} & \textbf{2.217} & 0.707 & \textbf{0.000} & 3.571 \\
& RtsGAN & 0.363 & 2.389 & 0.776 & \underline{1.325} & 18.951 & 12.926 & 5.464 & 0.384 & 0.081 & 4.000 \\
& SdeGAN & \textbf{0.267} & 3.659 & 0.813 & 1.542 & 42.017 & 38.541 & 65.557 & 0.390 & 0.979 & 5.143 \\
& LS4 & 0.298 & 6.041 & 0.855 & 2.457 & 40.362 & 24.262 & 69.751 & 0.797 & 0.805 & 6.143 \\
   \cmidrule[1pt]{2-12}
\end{tabular}
\label{tab:multisample_univariate_main_aggr}
\end{table}

\subsection{Utility-based metrics results}\label{sec:ut_met_res}

\begin{table}[t]
\caption{\textbf{Utility evaluation via fine-tuned forecasting models.} TTM forecasting performance on downstream tasks using different training sources: zero-shot, original data, generated data, and a combination of original and generated data. Results are reported for 3 multivariate datasets: \textit{bikesharing} (target: \texttt{count}, control: \texttt{temperature}, \texttt{humidity}), \textit{etth1} (target: \texttt{HUFL}, control: \texttt{MUFL}, \texttt{OT}), and \textit{traffic} (target: \texttt{junction1}, control: \texttt{junction2}, \texttt{junction3}). Metrics include RMSE, MASE, WQL, and average rank (lower is better). \textbf{Bold} highlights the best result within each row group; \underline{underlined} the second best; \underline{\textbf{bold+underlined}} the overall best.}
\vspace{0.1cm}
\centering
\tiny
\renewcommand{\arraystretch}{1.4}
\rowcolors{3}{gray!20}{white}
\begin{tabular}{>{\columncolor{white}[-20pt][\tabcolsep]}l lrrrrrrrrrr}
\cmidrule[1pt]{2-12}
& & \multicolumn{3}{c}{\textbf{bikesharing}} & \multicolumn{3}{c}{\textbf{etth1}} & \multicolumn{3}{c}{\textbf{traffic}} & \\
\cmidrule(lr){3-5} \cmidrule(lr){6-8} \cmidrule(lr){9-11} \rowcolor{white}
& & \textbf{RMSE} & \textbf{MASE} & \textbf{WQL} & \textbf{RMSE} & \textbf{MASE} & \textbf{WQL} & \textbf{RMSE} & \textbf{MASE} & \textbf{WQL} & \textbf{Avg. Rank} \\
\cmidrule[1pt]{2-12}
& 0-shot & 0.728 & 2.150 & 0.287 & 0.678 & 2.132 & 0.255 & 0.708 & \underline{\textbf{1.555}} & \textbf{0.255} & 1.78 \\
& Original Data (OD) & \textbf{0.495} & \textbf{0.822} & \textbf{0.178} & \textbf{0.658} & \textbf{1.820} & \textbf{0.232} & \textbf{0.702} & 1.995 & 0.283 & 1.22 \\
\cmidrule(lr){2-12}
\multirow{7}{*}{\rotatebox[origin=c]{90}{\textbf{GENERATED}}} & SDF-ICA & \textbf{0.514} & \underline{0.899} & \underline{0.194} & \textbf{0.626} & \textbf{1.820} & \underline{\textbf{0.224}} & 0.655 & 1.849 & \underline{0.262} & \textbf{2.00} \\
& SDF-FPC & 0.527 & 0.926 & 0.200 & 0.650 & 1.887 & 0.232 & 0.662 & 1.837 & 0.262 & 3.22 \\
& TimeVAE & 0.566 & 0.983 & 0.211 & 0.690 & 2.268 & 0.269 & 0.738 & 2.078 & 0.296 & 5.33 \\
& TimeVQVAE & \underline{0.520} & \textbf{0.867} & \textbf{0.188} & \textbf{0.626} & \underline{1.874} & \underline{0.227} & 0.702 & 1.995 & 0.283 & \underline{2.67} \\
& RtsGAN & 0.710 & 1.261 & 0.275 & 0.770 & 2.271 & 0.291 & \underline{\textbf{0.597}} & \textbf{1.574} & \underline{\textbf{0.225}} & 4.67 \\
& SdeGAN & 0.572 & 0.995 & 0.214 & 0.688 & 2.262 & 0.263 & \underline{0.629} & \underline{1.715} & \underline{0.243} & 4.00 \\
& LS4 & 0.839 & 1.468 & 0.318 & \underline{0.642} & 1.977 & 0.236 & 0.917 & 2.595 & 0.369 & 5.89 \\
\cmidrule(lr){2-12}
\multirow{7}{*}{\rotatebox[origin=c]{90}{\textbf{ORIGINAL + GEN}}} & SDF-ICA + OD & \underline{\textbf{0.487}} & \underline{\textbf{0.801}} & \underline{\textbf{0.173}} & 0.642 & \underline{\textbf{1.746}} & \underline{0.226} & 0.750 & 2.110 & 0.301 & \underline{3.22} \\
& SDF-FPC + OD & 0.493 & 0.829 & 0.179 & 0.666 & 1.754 & 0.231 & 0.743 & 2.087 & 0.297 & 4.78 \\
& TimeVAE + OD & \underline{0.492} & 0.814 & 0.176 & 0.654 & \underline{1.752} & 0.228 & 0.721 & 2.039 & 0.290 & \textbf{2.89} \\
& TimeVQVAE + OD & 0.495 & \underline{0.804} & \underline{0.174} & 0.678 & 1.887 & 0.242 & 0.724 & 2.043 & 0.291 & 4.56 \\
& RtsGAN + OD & 0.498 & 0.819 & 0.177 & \underline{0.637} & 1.872 & 0.231 & \underline{0.607} & \textbf{1.647} & \textbf{0.234} & 3.44 \\
& SdeGAN + OD & 0.495 & 0.837 & 0.181 & \underline{\textbf{0.620}} & 1.843 & \underline{\textbf{0.224}} & \textbf{0.605} & \underline{1.716} & \underline{0.242} & 3.33 \\
& LS4 + OD & 0.497 & 0.822 & 0.178 & 0.660 & 1.819 & 0.233 & 0.745 & 2.111 & 0.300 & 5.56 \\
\cmidrule[1pt]{2-12}
\end{tabular}
\label{tab:ttm_main}
\end{table}

Table~\ref{tab:ttm_main} presents the utility evaluation, where we assess the practical value of synthetic data by fine-tuning TTM on different multivariate training sources. 
For LS4, we modified the architecture to support multivariate generation. Furthermore, we adjust the embedding dimension $k$ in SDForger according to the complexity of each dataset, setting $k{=}3$ for \textit{bikesharing}, $k{=}7$ for \textit{etth1}, and $k{=}5$ for \textit{traffic}. To determine these values, we conducted a small ablation study to identify the optimal embedding dimension for each dataset (see Table~\ref{tab:ttm_appendix}).

SDForger demonstrates strong performance across datasets, as evidenced by its top average rank, with notable results on \textit{bikesharing} and \textit{etth1}. In \textit{bikesharing}, synthetic data from SDForger alone yields competitive scores, and combining it with real data leads to the best overall performance across metrics. On \textit{etth1}, SDForger-generated data surpasses original data in RMSE and WQL, suggesting it captures critical temporal and statistical structure. The hybrid setting (original + generated) maintains this advantage and further improves MASE. Performance on \textit{traffic} is more nuanced. Here, fine-tuning on real data is less effective, and GAN-based methods outperform others. Nevertheless, SDForger remains competitive, especially when using synthetic data alone. This suggests that the test distribution may deviate significantly from the training set, making traditional fine-tuning less useful. Indeed, high-quality synthetic data can act as a valuable supplement or even an alternative. 

In no scenario does synthetic data degrade downstream performance, underscoring the reliability and utility of SDForger-generated samples across varied forecasting contexts.

\subsection{Ablation}\label{sec:abl_c}

\textbf{Effect of embedding dimension $\bm{k}$}  
Appendix Table~\ref{tab:ablation_k_combined} presents an ablation study on the number of components $k$ used in SDForger’s embedding space. A compact embedding with $k{=}3$ offers strong performance across both \textit{multisample} and \textit{univariate} settings, indicating that a small number of components is often sufficient to capture core temporal and structural patterns. However, the optimal value of $k$ may vary across datasets, with more intricate dynamics potentially requiring higher-dimensional representations. In practice, users may also opt to select $k$ based on a desired percentage of explained variance, adapting the representation to specific application needs.

\textbf{Domain-level insights} 
Appendix
Tables~\ref{tab:multisample_per_data_average} and \ref{tab:univariate_per_data_average}
summarize the average normalized similarity scores per dataset. SDForger models achieve strong performance across a wide range of domains. Structured datasets such as \textit{Energy}, \textit{Appliances}, and \textit{Weather} exhibit particularly high scores. In contrast, domains such as \textit{Tourism}, \textit{Traffic}, and \textit{Finance} present greater challenges, likely due to their increased irregularity and noise. Nonetheless, SDForger maintains competitive results even in more complex settings, underscoring the flexibility of the proposed architecture.

\textbf{Generation efficiency}  
Appendix Table~\ref{tab:gen_time_by_length} reports the average time to generate univariate sequences for three targets from the \textit{Bikesharing} dataset, across two window lengths. SDForger is substantially faster than all competitors, often by one to two orders of magnitude. TimeVAE is the closest competitor but remains over 4$\times$ slower.  Notably, unlike GAN-based competitors, SDForger's generation time is independent of sequence length and scales with the number of embedding components ($k$). A minor exception occurs at $k=3$, where the reduced latent expressivity increases LLM fine-tuning time. Overall, SDForger achieves state-of-the-art efficiency without compromising quality.

\textbf{LLM comparison} \texttt{GPT-2} (124M) achieves performance on par with, and sometimes better than larger and more recent models such as \texttt{granite-3.0} (2B) and \texttt{phi-3.5} (3.8B) (Appendix Table~\ref{tab:ablation_llm_combined}). This shows that SDForger's pipeline is effective even leveraging
lightweight models. While runtime cost grows with model size, SDForger remains efficient compared to baselines (Appendix Table~\ref{tab:gen_time_llm}).

\textbf{Filtering Procedure.}
Appendix~Table~\ref{tab:filtering_rejection} reports rejection statistics for a representative generation scenario. 
We observe that the overall discard rate remains consistently low ($<2\%$) across settings, indicating that most generated embeddings fall within a plausible norm range. 
The proportion of missing values increases with larger embedding dimensions, reflecting a higher likelihood of incomplete generations in longer textual outputs.
Notably, the $\ell_2$ norms of accepted samples closely match those of the original embeddings, while discarded ones exhibit markedly higher values, confirming that the filtering procedure effectively removes divergent or anomalous generations.

\section{Shaping time series with language}\label{sec:TextConditional}

SDForger is designed to naturally incorporate textual information, making it well-suited for state-of-the-art time-series generation that embraces additional multi-modal inputs. To explore this, we conduct an experiment using the \textit{bikesharing} dataset. 
Coming from the intuition that these variables stem from a common physical process and may share latent components, we embed their three channels (temperature, count, and humidity) into a shared ICA basis. We incorporate the channel information in the textual encoder:
\texttt{``Condition:\ data is temp [sep] Input:\ value\_1 is [blank]...[sep]\ Target:} $e_{i1}$ \texttt{[answer]...''}

\begin{figure}[ht!]
  \centering
  \includegraphics[width=0.95\linewidth]{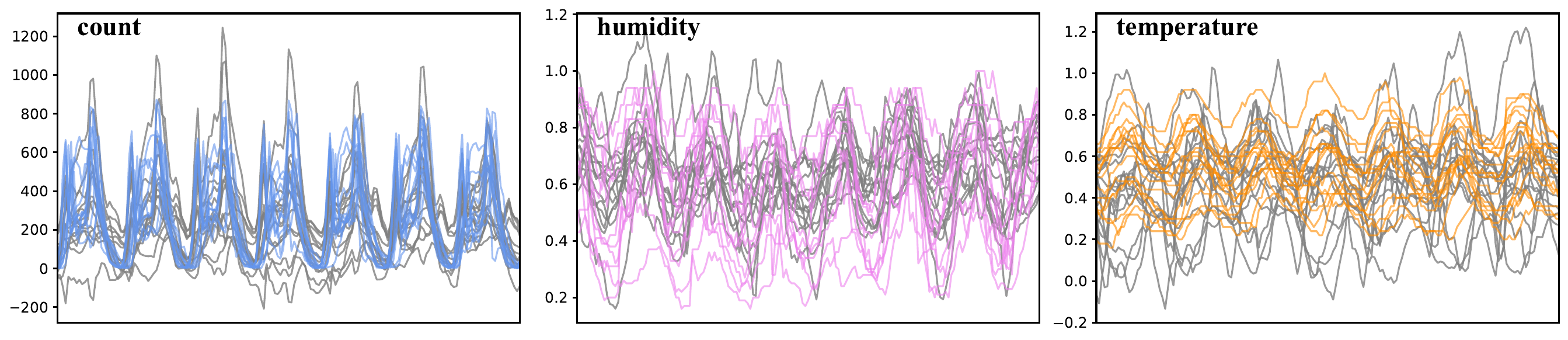} 
  \caption{\textbf{Text-Conditioned Generation with SDForger.} Visualization of 10 original (grey) and synthetic samples per channel from the \textit{bikesharing} data. Synthetic data is generated using conditional prompts: ``\texttt{Condition:\ data is cnt} (blue)'', ``\texttt{Condition:\ data is hum} (pink)'', and ``\texttt{Condition:\ data is temp} (orange)''.}
  \label{fig:textual_generated_curves}
\end{figure}

This conditioning strategy enables SDForger to generate channel-specific sequences with high fidelity. For instance, using a longitudinal $k$-nearest neighbor classifier \citep{ramos2024scikit} trained on real data, we achieve an accuracy of $0.81$ in identifying the generated curves (see Figure~\ref{fig:textual_generated_curves}). These results highlight SDForger’s strong generative capacity and its ability to integrate and respond to textual cues, positioning it as a flexible and powerful baseline for multimodal time-series synthesis.

%
%
\section{Conclusions}
\label{sec:Conclusions}

We introduced \textbf{SDForger}, a flexible and efficient framework for generating synthetic multivariate time series using large language models. By combining compact functional embeddings with textual conditioning, SDForger enables high-quality generation even in data-scarce settings. Extensive evaluations across multiple datasets and tasks demonstrate that SDForger consistently achieves strong similarity scores and enhances downstream forecasting performance—often matching or surpassing results obtained from real data and outperforming state-of-the-art baselines.

Ablation studies confirm the robustness of the framework across embedding strategies, dimensionality choices, and LLM architectures. SDForger is also highly efficient, with significantly lower generation times compared to its competitors. 
Moreover, by leveraging LLMs, SDForger enables seamless integration with textual prompts, paving the way for multimodal time-series generation, where natural language can guide not only content but also structure, semantics, or temporal context.

We believe SDForger can be further improved. Its modular design is intentionally built to support flexible experimentation, making it easy to explore enhancements or tailor components to specific needs. We see several promising directions:

\begin{itemize}[leftmargin=0.6cm, noitemsep, topsep=0pt]

\item \textbf{Embedding Strategies} While our current approach relies on linear methods like FastICA and FPC, future work could explore more expressive, nonlinear embeddings (e.g., AE) or multivariate-aware methods like Multivariate FPCA or Multivariate Singular Spectrum Analysis to better capture temporal and inter-channel dependencies.

\item \textbf{Parameter-Efficient Fine-Tuning} We currently use full fine-tuning for the LLM. However, using too many components relative to the number of instances can lead to unstable fine-tuning and reduced generation quality. Incorporating PEFT techniques such as LoRA or adapters could improve scalability, efficiency, and facilitate domain adaptation.

\item \textbf{Extension to encoder-only models} Our current implementation supports only autoregressive LLMs; future work would extend the framework to encoder-only models and different generation paradigms such as masked token prediction. 

\item \textbf{Extended Utility Evaluation} While we focus on forecasting, SDForger could be evaluated and optimized for broader downstream tasks such as classification or anomaly detection.

\item \textbf{Context and Covariate Integration} By design, SDForger supports integration of external covariates (e.g., categorical or textual data). Expanding this functionality could enable richer conditional generation, and multimodal transfer learning (see Section~\ref{sec:TextConditional}).

\end{itemize}

In summary, SDForger offers a flexible foundation, and we see meaningful opportunities to improve it both architecturally and in terms of task generalization.

\bibliographystyle{chicago}
\bibliography{bib}

%
%
\clearpage
\section*{Appendix}
\appendix 
\counterwithin{table}{section}
\counterwithin{figure}{section}

%
%

\section{Implementation details }
\label{appendix:ImplementationDetails}


\subsection{Data preprocessing: segmentation}
\label{appendix:data_preprocessing}

In many scenarios, each channel consists of a single historical time series, i.e., $I_0 = 1$. However, to estimate embeddings that effectively capture the temporal distribution, multiple instances per channel are necessary. Therefore, when $I_0 = 1$, we segment each channel into multiple overlapping windows. Specifically, for each channel $c$, we construct $I_c$ windows of fixed length $L_c$, where $L_c < L_0$. Without loss of generality, for simplicity in notation, we assume $L_c$ and $I_c$ are identical across all channels and denote them as $L$ and $I$, respectively.

To ensure robust learning of the embedding distribution, $I$ must be sufficiently large. Our experiments indicate that even $I = 15$ (i.e., 15 instances) suffices for this purpose. Once $I$ and $L$ are fixed, we segment the time series while minimizing the overlap between consecutive windows. The overlap step is determined by the dominant periodicity $P$ of the channel, ensuring that window transitions align with intrinsic temporal cycles.

The set of extracted windows $\mathcal{W}$ is formally defined as:
\begin{equation}
    \mathcal{W} = \{X[t:t+L] \mid t = 0, s, 2s, \dots, L_0 - L\}
\end{equation}
where the step size $s$ is computed as:
$s = \max(1, \lfloor \frac{L_0 - L}{I - 1} \rfloor)$
and then adjusted to be the nearest multiple of $P$ to maintain consistency in periodic structure.

To determine the dominant periodicity $P$, we employ the Autocorrelation Function (ACF), which quantifies the similarity between the time series and its lagged versions at different time shifts. This method is robust to noise and remains effective even when periodicity is not strictly stationary. 

The estimation of $P$ follows these steps:
\begin{enumerate}
    \item Compute the ACF and identify significant peaks, excluding lag $0$.
    \item Rank the detected peaks by their autocorrelation values.
    \item Select the highest-ranked period $P$ such that $P < L/2$.
\end{enumerate}

\begin{figure}[ht]
  \centering
  \includegraphics[width=1\linewidth]{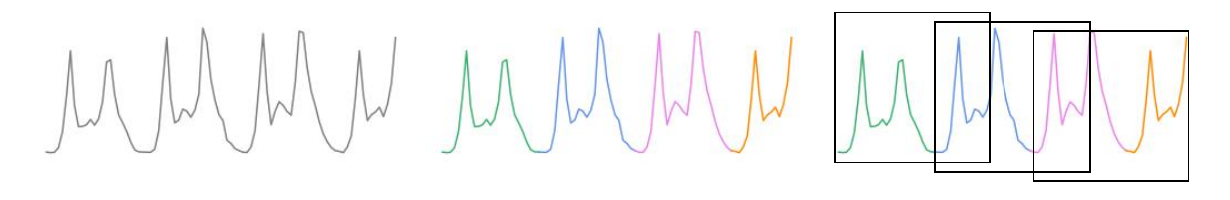} 
  \caption{\textbf{Periodicity-aware segmentation.}}
  \label{fig:segmentation}
\end{figure}

By leveraging this periodicity-aware segmentation strategy, we ensure that the extracted windows align with the natural cycles of the time series. Moreover, this approach minimizes window overlap, maximizing their independence and diversity and facilitating more effective embedding computations for downstream generative tasks.

This pre-processing transforms the data from a single sequence \( X_1 \) into a set \( X = \{X_i\}_{i=1}^I \), where each \( X_i \in \mathbb{R}^{C \times L} \), preparing the data for the generation task.


\subsection{Choice of the number of components}
\label{appendix:k_selection}

The choice of \( k_c \) determines how well the basis representation approximates the original time-series channel \( c \). Our frameword allows the user to either select the smallest \( k_c \) that explains a predefined percentage of the total variance of the original time series or manually specify \( k_c \).

There exists an inherent trade-off in selecting \( k_c \). A higher \( k_c \) captures more of the total variability but can hinder the LLM’s ability to model the underlying distribution during the generation phase. Moreover, if \( k_c \) is too large, the generated samples may become overly similar to the original data, limiting diversity and the introduction of novel patterns. Conversely, choosing \( k_c \) too small risks omitting essential structures and temporal characteristics, degrading the reconstruction quality.

It is important to note that the nature of the components differs between FPC and FastICA. FPC forms a parsimonious basis system, where a few components typically suffice to capture most of the variability, with components ordered by the amount of variance they explain—early components being systematically more informative. In contrast, FastICA components are unordered: each component contributes independently, without a hierarchical importance structure. As a result, FastICA generally requires more components to achieve a similar reconstruction quality compared to FPC. However, this property also makes FastICA embeddings more robust during generation, as information is distributed more evenly across components, reducing the risk that a few badly generated components disproportionately affect the synthesized curves.

To provide a quantitative intuition, Appendix~Table~\ref{tab:variance_fica_fpc} reports the proportion of variance retained across embedding dimensions for both decomposition methods. 
This analysis highlights how the variance explained increases with \(k\), and how FPC typically achieves higher cumulative variance with fewer components, while FastICA distributes information more evenly across dimensions.

In practice, we recommend keeping the total number of components \( K \) reasonably small, particularly when the number of training instances is limited. Empirically, with a training set of 30 instances, setting \( K > 25 \) often results in unstable fine-tuning and an increased rate of discarded samples due to low-quality generation. This limitation stems from the LLM's reduced ability to model high-dimensional embeddings under data-scarce conditions effectively.



\subsection{In-generation filtering}
\label{appendix:filtering}

\paragraph{Missing values and duplicated instances.} 
To illustrate the filtering logic, we report three concrete examples of generated prompts with embedding dimension $K = 4$. 
Following the inference template:
\begin{align*}
\mathcal{P}_g^{\textup{INF}} = \texttt{``Input:} 
\bigcirc_{k=1}^{K} \texttt{( value\_{$\pi(k)$} is [blank], )} 
\texttt{ [sep] Target:''},
\end{align*}
the model produces the following textual generations:

\begin{itemize}[leftmargin=0.6cm, noitemsep, topsep=0pt]
    \item \textbf{Prompt 1 (valid)}\\
    \texttt{Input: value\_2 is [blank], value\_4 is [blank], value\_1 is [blank], value\_3 is [blank] [sep] Target: 0.125 [answer] -0.084 [answer] 0.217 [answer] 0.041 [answer]}\\
    \item \textbf{Prompt 2 (duplicated)}\\
    \texttt{Input: value\_4 is [blank], value\_1 is [blank], value\_2 is [blank], value\_3 is [blank] [sep] Target:  -0.084 [answer] 0.217 [answer] 0.125 [answer] 0.041 [answer]}\\
    \item \textbf{Prompt 3 (missing value)}\\
    \texttt{Input: value\_1 is [blank], value\_3 is [blank], value\_2 is [blank], value\_4 is [blank] [sep] Target: 0.182 [answer] 0.095 [answer] -0.012 [answer]}
\end{itemize}

In this example, the filtering stage identifies \emph{Prompt 2} as a duplicate of \emph{Prompt 1} and discards it, while \emph{Prompt 3} is removed because it does not contain all the targets' coefficients.
Only \emph{Prompt 1} is retained for reconstruction. 
This simple yet effective procedure ensures that the generated embedding tables remain diverse and valid before decoding into the time-series domain.

\paragraph{Diverging instances} To ensure the quality of synthetic data, we discard generated instances whose embedding coefficients significantly deviate from the distribution of the original data. Specifically, we compute the squared $\ell_2$-norm of each embedding vector and compare it to the norms of the original embeddings. This criterion efficiently filters out extreme outliers in the latent space, without requiring reconstruction into the time-series domain.

Formally, for each channel \( c \), let \( \hat{E}^{c,s} = E^c \cup \tilde{E}^{c,\leq s-1} \) be the matrix containing both original embeddings and all previously accepted generated embeddings up to inference step \( s-1 \), with \( \hat{E}^{c,0} = E^c \). Denote by \( N^{c,\text{old}} \) and \( N^{c,\text{new}} \) the sets of squared Euclidean norms of the rows of \( \hat{E}^{c,s-1} \) and the newly generated matrix \( \tilde{E}^{c,s} \), respectively. For each newly generated row \( i \), we compute its norm and accept it only if:

\begin{equation*}
    q_1 - 3 \cdot \text{IQR} \leq N^{c,\text{new}}_i \leq q_3 + 3 \cdot \text{IQR},
\end{equation*}

where \( q_1 \) and \( q_3 \) are the first and third quartiles of \( N^{c,\text{old}} \), and \( \text{IQR} = q_3 - q_1 \). An instance is retained only if this condition is satisfied across all channels \( c \).

This norm-based strategy is particularly well-motivated when using FPCs, due to the orthonormality of the basis. Let \( \mathcal{X}^c_i \) be a time series in channel \( c \) and \( b^c_j \) the corresponding FPC basis. Then the $\mathbb{L}^2$-norm of \( \mathcal{X}^c_i \) can be approximated by the Euclidean norm of its FPC coefficients:

\begin{equation*}
    \int_\mathcal{T} (\mathcal{X}^c_i)^2 \, \mathrm{d}t = \lVert \mathcal{X}^c_i \rVert_{\mathbb{L}^2}^2 = 
    \sum_{j=1}^{\infty} \langle \mathcal{X}^c_i, b^c_j \rangle_{\mathbb{L}^2}^2 \approx \sum_{j=1}^{k_c} (e^c_{ij})^2,
\end{equation*}

where \( e^c_{ij} \) are the FPC embedding coefficients. This justifies the norm-based filtering as a direct proxy for detecting time-series samples with unusually high or low energy.


While filtering based on coefficient norms does not guarantee full statistical fidelity to the original time-series distribution, it serves as an effective mechanism to remove extreme outliers without reconstruction. Combined with additional checks for missing values and duplicates, this step helps preserve both the \emph{diversity} and \emph{relevance} of the generated data. A high rejection rate may indicate insufficient LLM fine-tuning or poor generalization, suggesting the need for more representative training data or additional training steps.


\subsection{Stopping criterion}
\label{appendix:stopping}

The stopping criterion monitors the diversity of the generated norms across all channels to determine when the generation process should stop. 
When using FPC, these norms correspond to the \( \mathbb{L}^2 \)-norms of the generated curves. When using FastICA, they correspond to the norms of the embedding coefficient vectors; although not directly related to the curve norms, they still provide a useful proxy for identifying over-sampling and loss of variability.

At inference step \( s \), for each channel \( c \), let \( u^c \) denote the number of unique values in \( N^{c,\text{old}} \) (the set of accepted norms up to step \( s \)), rounded to the fourth decimal place. Let \( \tilde{I} \) be the total number of valid instances generated so far. We define the \emph{diversity score} for channel \( c \) as:
\[
D^c = {u^c}/{\tilde{I}}.
\]

The diversity score provides a quantitative measure of how much variability remains in the generated norms. We track \( D^c \) at each inference step, and the stopping condition is triggered when:

\[
\max_{c} D^c < \lambda_{\text{stop}} \quad \text{or} \quad \tilde{I} > \tilde{I}_{\max}.
\]

In other words, generation stops either when the maximum diversity score across channels falls below a predefined threshold \( \lambda_{\text{stop}} \), indicating reduced novelty, or when the total number of generated instances exceeds a maximum cap \( \tilde{I}_{\max} \).

Monitoring the diversity score enables us to assess whether the model continues to introduce new variability in the generated data, serving as an online signal for generation quality.

If we denote by \( S \) the final inference step, then the output of the generation process is the complete embedding table \( \tilde{E} \in \mathbb{R}^{\tilde{I} \times K} \), where, for each channel \( c \), \( \tilde{E}^c = \tilde{E}^{c, \leq S} \).



%
%
\section{Evaluation metrics}
\label{appendix:evaluation_metrics}

All the metrics presented below are adopted from~\cite{ang2023tsgbench}, except for \textit{Shapelet-based Reconstruction}, which follows the definition in~\cite{zheng2016efficient}.

\subsection{Feature-based evaluation 
}

\paragraph{Marginal Distribution Difference} MDD computes an empirical histogram for each dimension and time step in the generated series, using the bin centers and widths from the original series. It then calculates the average absolute difference between this histogram and that of the original series across bins, assessing how closely the distributions of the original and generated series align.

\paragraph{AutoCorrelation Difference} ACD computes the autocorrelation of both the original and generated time series, then determines their difference. By contrasting the autocorrelations, we could evaluate how well dependencies are maintained in the generated time series.

\paragraph{Skewness Difference} SD is vital for the marginal distribution of a time series, quantifying its distribution asymmetry. Given the mean (standard deviation) of the train time series $T_s^{tr}$ as $\mu_s^{tr}$ ($\sigma_s^{tr}$) and the generated time series $T_s^{gen}$ as $\mu_s^{gen}$ ($\sigma_s^{gen}$), we evaluate the fidelity of $T_s^{gen}$ by computing the skewness difference between them as:
\begin{equation*}
        SD = \left| \frac{\mathbb{E}[(T_s^{gen} - \mu_s^{gen})^3]}{\sigma_s^{gen3}} - \frac{\mathbb{E}[(T_s^{tr} - \mu_s^{tr})^3]}{\sigma_s^{tr3}} \right|.
\end{equation*}

\paragraph{Kurtosis Difference} Like skewness, KD assesses the tail behavior of a distribution, revealing extreme deviations from the mean. Using the previous notations, the kurtosis difference between $T_s^{tr}$ and $T_s^{gen}$ is calculated as:
\begin{equation*}
        KD = \left| \frac{\mathbb{E}[(T_s^{gen} - \mu_s^{gen})^4]}{\sigma_s^{gen4}} - \frac{\mathbb{E}[(T_s^{tr} - \mu_s^{tr})^4]}{\sigma_s^{tr4}} \right|.
\end{equation*}

\subsection{Distance-based evaluation 
}

\paragraph{Euclidean Distance} For each original series $s^{tr} = (x_1,...,x_l)$ and its generated $s^{gen} = (y_1,...,y_l)$, $ ED = \sqrt{ \sum_{1=1}^{l} (x_i-y_i)^2 }$. We take the mean of ED for all series and all samples. Given that the input time series has been preprocessed to fit within the range of $[0, 1]$, ED deterministically
assesses the similarity between $s^{gen}$ and $s^{tr}$. It provides a value-wise comparison between the time series.

\paragraph{Dynamic Time Warping} Given that ED overlooks alignment, we include DTW to capture the optimal alignment between series regardless of their pace or timing.
The alignment facilitated by DTW offers insights into the predictive quality of the generated series. 

\paragraph{Shapelet-based Reconstructions} Shapelet based RE is calculated by generated time series using shapelets extracted using shift invariant dictionary learning (SIDL) algorithm \citep{zheng2016efficient}. Shapelets represent local discriminative patterns present in the time-series data. We learn shift invariant patterns/shapelets on the original time-series dataset and then use the learnt dictionary to reconstruct unseen generated time series. The reconstruction error is calculated between the generated time series and their reconstruction using SIDL.







%
%

\section{Datasets \& protocol settings}
\label{appendix:datasets}

\begin{table}[H]
\caption{\textbf{Overview of the benchmark datasets.} For each dataset, we report its application domain, sampling frequency, number of time series, length statistics, and the type of evaluation (univariate, multivariate, multi-sample) it supports.}
\vspace{0.1cm}
\centering
\tiny
\renewcommand{\arraystretch}{1.3} 
\rowcolors{4}{gray!20}{white}
\setlength{\tabcolsep}{6pt} 
\begin{tabular}{llrrccc}
\toprule
\textbf{Dataset} &
\textbf{Domain} &
\textbf{Freq.} &
\textbf{Number} &
\multicolumn{3}{c}{\textbf{Evaluation type}} \\
\cmidrule(lr){5-7} 
& & & & \textbf{MS} & \textbf{UV} & \textbf{MV}\\
\midrule
{Australian Electricity} & energy & 30m & 5 & N & Y & Y \\
{Appliances} & energy & 10m & 1 & N & Y & N \\
{Bikesharing} & general & 1H & 3 & N & Y & Y \\
{Carbon Capture Plant} & nature & 2m & 4 & N & Y & N \\
{ETTH1} & energy & 1H & 3 & N & Y & Y \\
{ECL} & energy & 1H & 320  & Y & N & N \\
{Exchange Rate} & finance & 1D & 8 & N & Y & N \\
{NN5} & finance & 1D & 111 & Y & N & N \\
{Tourism} & general & 1M & 365 & Y & N & N \\
{Traffic} & transport & 1H & 3 & N & Y & Y \\
{Traffic Monash} & transport & 1H & 861 & Y & N & N \\
{Solar - Weather} & nature & 1H & 653 & Y & N & N \\
{Rain - Weather} & nature & 1H & 386 & Y & N & N \\
{Temperature - Weather} & nature & 1H & 362 & Y & N & N \\
\bottomrule
\end{tabular}
\label{tab:dataset_descritpion}
\end{table}

\subsection{Dataset overview}


\paragraph{Energy}
\begin{itemize}[leftmargin=0.6cm, noitemsep, topsep=0pt]
    \item \emph{Australian Electricity} \citep{godahewa2021monash} contains electricity demand data from 5 states in Australia.
    \item \emph{Appliances}\footnote{\url{https://www.kaggle.com/datasets/loveall/appliances-energy-prediction}} contains house temperature and humidity conditions monitored with a wireless sensor network, and energy data logegd with m-bus energy meters averaged for 10 minutes periods.
    \item \emph{ETTH1}\footnote{\url{https://github.com/zhouhaoyi/ETDataset}} contains oil temperatures and other covariates of electrical transformers from two stations in China, measured at 15 minutes granularity but hourly aggregated.
    \item \emph{ECL} \footnote{\url{https://www.kaggle.com/datasets/minhnguyendichnhat/ecl-dataset}} contains electricity consumption of 370 points.
\end{itemize}

\paragraph{Mobility and Transport}
\begin{itemize}[leftmargin=0.6cm, noitemsep, topsep=0pt]
    \item \emph{Bikesharing}\footnote{\url{https://www.kaggle.com/datasets/lakshmi25npathi/bike-sharing-dataset}} contains the hourly and daily count of rental bikes between the years 2011 and 2012 in the Capital bike share system with the corresponding weather and seasonal information.
    \item \emph{Traffic}\footnote{\url{https://www.kaggle.com/datasets/fedesoriano/traffic-prediction-dataset}} contains observations of the number of vehicles each hour in four different junctions
    \item \emph{Traffic Monash} \citep{godahewa2021monash} contains hourly road occupancy readings from sensors in the San Francisco Bay area.
    \item \emph{Tourism} \citep{godahewa2021monash} dataset from, used for the Kaggle Tourism Forecasting competition. This dataset is non-stationary.
\end{itemize}

\paragraph{Nature}
\begin{itemize}[leftmargin=0.6cm, noitemsep, topsep=0pt]
    \item \emph{Carbon Capture Plant} \citep{jablonka2023machine} records the emission profiles of “2-amino-2-methyl-1-propanol” (AMP) and “piperazine” (Pz) collected at every 2 minutes interval.
    \item \emph{Weather} \citep{godahewa2021monash} contains daily time series of four weather variables (rain, mintemp, maxtemp and solar radiation) measured at weather stations in Australia.
\end{itemize}

\paragraph{Finance}
\begin{itemize}[leftmargin=0.6cm, noitemsep, topsep=0pt]
    \item \emph{Exchange Rate} \citep{godahewa2021monash} contains daily exchange rates for currencies of eight countries (Australia, British, Canada, Switzerland, China, Japan, New Zealand and Singapore) between 1990 and 2016. This dataset is non-stationary.
    \item \emph{NN5 (Daily, Weekly)} \citep{godahewa2021monash} contains cash withdrawal data from ATMs. This dataset combines stationary and non-stationary time series.
\end{itemize}

\subsection{Protocols}

\paragraph{Multisample setting} For MS data preparation, we sampled $I=30$ instances, each of length $L=250$, resulting in a total training sequence of 15,000 timestamps. Standard scaling per timestamp is applied. For evaluation, we generate 100 synthetic instances. 

\paragraph{Univariate setting} For UV data preparation, we used a training sequence of length $L_0=2,000$, segmented into $I=30$ instances of length $L=250$ using our periodicity-aware segmentation strategy (cf Appendix~\ref{appendix:data_preprocessing}). Same standard scaling was applied. For evaluation, we generate 100 synthetic instances. 

\paragraph{Multivariate setting}
For fine-tuning TTM in the MV setting, we used consecutive sequences of length $L_0 = 5000$, $2500$, and $2500$ for training, validation, and testing, respectively. Each set was segmented into $I = 30$, $15$, and $15$ instances of length $L = 1120$ using our period-aware segmentation strategy (Appendix~\ref{appendix:data_preprocessing}). 
Periodicity was estimated from the training set. Standard scaling per timestamp, computed on the training set, was applied consistently across all splits. For evaluation, we generated 30 instances, matching the size of the training set.





%
%
\section{Additional results}
\label{appendix:additional_results}

\begin{table}[H]
\caption{\textbf{Variance retained across embedding dimensions.} 
For each dataset, we report the proportion of total variance retained for embedding dimensions \(k = 3\), \(k = 5\), and \(k = 7\) under both FastICA and FPC decompositions. 
}
\vspace{0.1cm}
\centering
\tiny
\renewcommand{\arraystretch}{1.3} 
\rowcolors{4}{gray!20}{white}
\setlength{\tabcolsep}{4pt} 
\begin{tabular}{lcccccc}
\toprule
\textbf{Dataset} &
\multicolumn{3}{c}{\textbf{FICA}} &
\multicolumn{3}{c}{\textbf{FPC}} \\
\cmidrule(lr){2-4} \cmidrule(lr){5-7} 
& \textbf{k=3} & \textbf{k=5} & \textbf{k=7} & \textbf{k=3} & \textbf{k=5} & \textbf{k=7} \\
\midrule
Appliances & 0.333 & 0.457 & 0.539 & 0.324 & 0.459 & 0.555 \\
Australian Electricity & 0.703 & 0.853 & 0.912 & 0.787 & 0.895 & 0.938 \\
Bikesharing & 0.492 & 0.662 & 0.744 & 0.599 & 0.734 & 0.796 \\
Carbon Capture Plant & 0.773 & 0.905 & 0.951 & 0.843 & 0.933 & 0.965 \\
ETTH1 & 0.536 & 0.686 & 0.775 & 0.635 & 0.754 & 0.823 \\
ECL & 0.809 & 0.941 & 0.971 & 0.991 & 0.997 & 0.999 \\
Exchange Rate & 0.776 & 0.907 & 0.942 & 0.957 & 0.982 & 0.989 \\
NN5 & 0.336 & 0.479 & 0.590 & 0.719 & 0.780 & 0.826 \\
Tourism & 0.869 & 0.954 & 0.977 & 0.986 & 0.995 & 0.998 \\
Traffic & 0.376 & 0.564 & 0.697 & 0.408 & 0.591 & 0.714 \\
Traffic Monash & 0.541 & 0.722 & 0.823 & 0.745 & 0.845 & 0.902 \\
Rain - Weather & 0.449 & 0.605 & 0.725 & 0.558 & 0.684 & 0.781 \\
Solar - Weather & 0.450 & 0.616 & 0.718 & 0.674 & 0.773 & 0.833 \\
Temperature Max - Weather & 0.493 & 0.602 & 0.685 & 0.864 & 0.893 & 0.915 \\
Temperature Min - Weather & 0.390 & 0.517 & 0.611 & 0.777 & 0.824 & 0.858 \\
\bottomrule
\end{tabular}
\label{tab:variance_fica_fpc}
\end{table}

\begin{table}[H]
\caption{
\textbf{Filtering statistics for generated embeddings.} 
Rejection statistics on the \textit{count} variable from the \textit{bikesharing} dataset, averaged across 5 seeds.
Each row reports the proportion of generated samples containing missing values, the fraction of samples discarded by the filtering stage, and the average $\ell_2$ norms of the original, accepted, and discarded embedding vectors.}
\vspace{0.1cm}
\centering
\tiny
\renewcommand{\arraystretch}{1.4}
\rowcolors{2}{gray!20}{white}
\begin{tabular}{lrrrrr}
    \cmidrule[1pt]{1-6}
    & \textbf{NaN\%} & \textbf{Discard\%} & \textbf{Norms Original (Avg)} & \textbf{Norms Accepted (Avg)} & \textbf{Norms Discarded (Avg)} \\
    \cmidrule[1pt]{1-6}
    SDF-ICA$_3$ & 3.87 & 1.94 & 1.708 & 1.714 & 19.066 \\
    SDF-ICA$_5$ & 36.29 & 1.94 & 2.202 & 2.248 & 9.514 \\
    SDF-ICA$_7$ & 58.82 & 0.00 & 2.602 & 2.521 & 0.000 \\
    \cmidrule[1pt]{1-6}
\end{tabular}
\label{tab:filtering_rejection}
\end{table}

\begin{table}[H]
\caption{\textbf{Per-dataset similarity results in the multisample setting.} Average normalized similarity scores (feature-based and distance-based) for each dataset and model.}
\vspace{0.1cm}
\centering
\tiny
\renewcommand{\arraystretch}{1.4}
\rowcolors{2}{gray!20}{white}
\begin{tabular}{lcccccccccc}
    \toprule
    & \multicolumn{5}{c}{\textbf{Feature-Based}} & \multicolumn{5}{c}{\textbf{Distance-Based}} \\
    \cmidrule(lr){2-6} \cmidrule(lr){7-11}
    & \textbf{ecl} & \textbf{nn5} & \textbf{tourism} & \textbf{traffic} & \textbf{weather} 
    & \textbf{ecl} & \textbf{nn5} & \textbf{tourism} & \textbf{traffic} & \textbf{weather} \\
    \midrule
SDF-ICA$_3$ & 0.402 & \underline{0.195} & 0.268 & 0.330 & \underline{0.204} & \underline{0.018} & 0.129 & \underline{0.032} & 0.086 & 0.137 \\
SDF-FPC$_3$ & 0.576 & 0.323 & 0.594 & \underline{0.293} & 0.296 & 0.054 & 0.183 & 0.073 & 0.099 & 0.136 \\
TimeVAE & \underline{0.164} & \textbf{0.109} & \underline{0.211} & \textbf{0.143} & \textbf{0.174} & 0.063 & 0.139 & 0.074 & 0.143 & 0.125 \\
TimeVQVAE & 0.754 & 0.380 & 0.818 & 0.437 & 0.498 & \textbf{0.003} & \textbf{0.068} & \textbf{0.009} & \textbf{0.053} & \textbf{0.069} \\
RtsGAN & \textbf{0.049} & 0.347 & \textbf{0.174} & 0.344 & 0.315 & 0.075 & \underline{0.074} & 0.110 & \underline{0.059} & \underline{0.102} \\
SdeGAN & 0.499 & 0.219 & 0.539 & 0.391 & 0.308 & 0.299 & 0.652 & 0.290 & 0.496 & 0.616 \\
LS4 & 0.860 & 0.319 & 0.849 & 0.470 & 0.400 & 0.720 & 0.681 & 0.639 & 0.926 & 0.707 \\
    \bottomrule
\end{tabular}
\label{tab:multisample_per_data_average}
\end{table}

\begin{table}[H]
\caption{\textbf{Per-dataset similarity results in the univariate setting.} Average normalized similarity scores (feature-based and distance-based) for each dataset and model.}
\vspace{0.1cm}
\hspace{-0.3cm}
\tiny
\renewcommand{\arraystretch}{1.4}
\rowcolors{2}{gray!20}{white}
\begin{tabular}{lcccccccccccccc}
\toprule
& \multicolumn{7}{c}{\textbf{Feature-Based}} & \multicolumn{7}{c}{\textbf{Distance-Based}} \\
\cmidrule(lr){2-8} \cmidrule(lr){9-15}
& \textbf{appl.} & \textbf{austr.} & \textbf{bike} & \textbf{carbon} & \textbf{etth1} & \textbf{exch.} & \textbf{traffic}
& \textbf{appl.} & \textbf{austr.} & \textbf{bike} & \textbf{carbon} & \textbf{etth1} & \textbf{exch.} & \textbf{traffic} \\
\midrule
SDF-ICA$_3$ & \underline{0.486} & \textbf{0.122} & \textbf{0.162} & \underline{0.197} & \underline{0.128} & 0.155 & 0.269 & 0.088 & \underline{0.073} & \underline{0.077} & \underline{0.063} & 0.080 & 0.115 & 0.059 \\
SDF-FPC$_3$ & 0.509 & \textbf{0.122} & \underline{0.169} & 0.345 & \textbf{0.107} & 0.287 & \textbf{0.213} & 0.085 & 0.101 & \underline{0.077} & 0.078 & 0.076 & 0.143 & 0.064 \\
TimeVAE & \textbf{0.370} & 0.123 & 0.184 & \textbf{0.176} & 0.234 & \textbf{0.113} & \underline{0.236} & 0.119 & 0.079 & 0.155 & 0.102 & 0.140 & \underline{0.091} & 0.178 \\
TimeVQVAE & 0.531 & 0.439 & 0.371 & 0.572 & 0.354 & 0.578 & 0.312 & \textbf{0.045} & \textbf{0.031} & \textbf{0.037} & \textbf{0.005} & \textbf{0.031} & \textbf{0.031} & \textbf{0.035} \\
RtsGAN & 0.567 & 0.300 & 0.273 & 0.226 & 0.244 & 0.217 & 0.312 & \underline{0.056} & 0.098 & 0.084 & 0.085 & \underline{0.072} & 0.157 & \underline{0.053} \\
SdeGAN & 0.634 & 0.128 & 0.174 & 0.409 & 0.173 & \underline{0.115} & 0.249 & 0.456 & 0.623 & 0.831 & 0.703 & 0.833 & 0.469 & 0.765 \\
LS4 & 0.609 & 0.294 & 0.265 & 0.695 & 0.289 & 0.416 & 0.300 & 0.720 & 0.471 & 0.475 & 0.554 & 0.364 & 0.498 & 0.621 \\
\bottomrule
\end{tabular}
\label{tab:univariate_per_data_average}
\end{table}

\begin{table}[H]
\caption{
\textbf{Average generation time: baselines} Average time (in seconds) required to generate synthetic univariate time series for the \texttt{bikesharing} dataset across three targets: \texttt{count}, \texttt{temperature}, and \texttt{humidity}. We report results for two input sequence lengths: 250 and 500. All models were evaluated under the same computational constraints (\texttt{-mem 20G -cores 1+1 -gpu v100}) using a single NVIDIA V100 GPU.}
\vspace{0.1cm}
\tiny
\renewcommand{\arraystretch}{1.4}
\rowcolors{2}{gray!20}{white}
\hspace*{-1cm}
\begin{tabular}{l|rrrrrrrrrrr}
    \toprule
    \textbf{Length} & SDF-ICA$_3$ & SDF-ICA$_5$ & SDF-ICA$_7$ & SDF-FPC$_{3}$ & SDF-FPC$_{5}$ & SDF-FPC$_{7}$ & TimeVAE & TimeVQVAE & RtsGAN & SdeGAN & LS4 \\
    \midrule
    \textbf{250} & 41.9 & 26.8 & 28.8 & 22.0 & 25.4 & 33.1 & 138.1 & 4574.2 & 2055.7 & 3498.9 & 2804.4 \\
    \textbf{500} & 38.3 & 22.8 & 26.0 & 17.9 & 22.8 & 26.6 & 112.6 & 4401.4 & 3536.3 & 7316.7 & 2378.9 \\
    \bottomrule
\end{tabular}
\label{tab:gen_time_by_length}
\end{table}

\begin{table}[H]
\caption{\textbf{Ablation study: embedding dimension.} Aggregated similarity-based performance across all datasets in the \textbf{multisample} and \textbf{univariate} setting.}
\vspace{0.1cm}
\centering
\tiny
\renewcommand{\arraystretch}{1.4}
\rowcolors{2}{gray!20}{white}
\begin{tabular}{>{\columncolor{white}[-20pt][\tabcolsep]}l lrrrrrrrrr}
    \cmidrule[1pt]{2-11}
    & & \multicolumn{4}{c}{\textbf{Feature-based}} & \multicolumn{3}{c}{\textbf{Distance-based}} & \multicolumn{2}{c}{\textbf{Norm. Avg.}} \\
    \cmidrule(lr){3-6} \cmidrule(lr){7-9} \cmidrule(lr){10-11}
    & & \textbf{MDD} & \textbf{ACD} & \textbf{SD} & \textbf{KD} & \textbf{ED} & \textbf{DTW} & \textbf{SHR} & \textbf{Feat.} & \textbf{Dist.} \\
    \cmidrule{2-11}
\multirow{6}{*}{\rotatebox[origin=c]{90}{\textbf{MULTISAMPLE}}}
& SDF-FPC$_{3}$ & \underline{0.255} & 2.166 & \underline{1.323} & 4.299 & 17.749 & 11.921 & 16.537 & 0.616 & 0.609 \\
& SDF-FPC$_{5}$ & 0.262 & 3.191 & 1.336 & 3.668 & 17.475 & \underline{11.727} & 22.893 & 0.714 & 0.535 \\
& SDF-FPC$_{7}$ & 0.264 & 3.534 & 1.500 & 3.560 & 17.710 & \textbf{11.652} & 28.068 & 0.787 & 0.655 \\
& SDF-ICA$_{3}$ & \textbf{0.244} & 1.180 & \textbf{0.869} & \textbf{2.384} & \textbf{16.669} & 12.373 & \textbf{6.870} & \textbf{0.050} & \underline{0.333} \\
& SDF-ICA$_{5}$ & 0.261 & \underline{0.782} & 1.378 & \underline{2.649} & \underline{16.743} & 12.238 & \underline{7.731} & \underline{0.371} & \textbf{0.307} \\
& SDF-ICA$_{7}$ & 0.265 & \textbf{0.589} & 1.964 & 2.963 & 16.900 & 12.031 & 14.195 & 0.576 & 0.362 \\
\addlinespace[0.5em]
\cmidrule{2-11}
\addlinespace[0.5em]
\multirow{6}{*}{\rotatebox[origin=c]{90}{\textbf{UNIVARIATE}}}
& SDF-FPC$_{3}$ & 0.308 & 1.480 & 0.801 & 1.690 & 19.340 & 12.809 & \underline{5.452} & 0.736 & 0.469 \\
& SDF-FPC$_{5}$ & 0.306 & 1.887 & 0.773 & 1.581 & 20.534 & 12.513 & 8.920 & 0.536 & 0.753 \\
& SDF-FPC$_{7}$ & 0.309 & 2.399 & 0.774 & 1.954 & 20.470 & \textbf{12.079} & 10.982 & 0.947 & 0.654 \\
& SDF-ICA$_{3}$ & \underline{0.306} & 1.396 & \textbf{0.671} & \underline{1.382} & \textbf{18.802} & 12.435 & \textbf{4.856} & \textbf{0.169} & \textbf{0.163} \\
& SDF-ICA$_{5}$ & 0.306 & \underline{0.867} & 0.770 & \textbf{1.333} & \underline{19.043} & \underline{12.261} & 6.555 & 0.279 & \underline{0.222} \\
& SDF-ICA$_{7}$ & \textbf{0.306} & \textbf{0.597} & \underline{0.736} & 1.458 & 19.989 & 12.381 & 8.102 & \underline{0.175} & 0.543 \\
    \cmidrule[1pt]{2-11}
\end{tabular}
\label{tab:ablation_k_combined}
\end{table}

\begin{table}[H]
\caption{\textbf{Ablation study: embedding dimension.} TTM forecasting performance on downstream tasks using different training sources: generated data, and a combination of original and generated data. Results are reported for 3 multivariate datasets: \textit{bikesharing} (target: \texttt{count}, control: \texttt{temperature}, \texttt{humidity}), \textit{etth1} (target: \texttt{HUFL}, control: \texttt{MUFL}, \texttt{OT}), and \textit{traffic} (target: \texttt{junction1}, control: \texttt{junction2}, \texttt{junction3}). Metrics include RMSE, MASE, WQL, and average rank (lower is better). \textbf{Bold} highlights the best result within each row group; \underline{\textbf{bold+underlined}} the overall best.}
\vspace{0.1cm}
\centering
\tiny
\renewcommand{\arraystretch}{1.4}
\rowcolors{3}{gray!20}{white}
\begin{tabular}{>{\columncolor{white}[-20pt][\tabcolsep]}l lrrrrrrrrrr}
\cmidrule[1pt]{2-11}
& & \multicolumn{3}{c}{\textbf{bikesharing}} & \multicolumn{3}{c}{\textbf{etth1}} & \multicolumn{3}{c}{\textbf{traffic}} & \\
\cmidrule(lr){3-5} \cmidrule(lr){6-8} \cmidrule(lr){9-11} \rowcolor{white}
& & \textbf{RMSE} & \textbf{MASE} & \textbf{WQL} & \textbf{RMSE} & \textbf{MASE} & \textbf{WQL} & \textbf{RMSE} & \textbf{MASE} & \textbf{WQL} \\
\cmidrule[1pt]{2-11}
& 0-shot & 0.728 & 2.150 & 0.287 & 0.678 & 2.132 & 0.255 & 0.708 & \underline{\textbf{1.555}} & \underline{\textbf{0.255}} \\
& Original Data (OD) & \textbf{0.495} & \textbf{0.822} & \textbf{0.178} & \textbf{0.658} & \textbf{1.820} & \textbf{0.232} & \textbf{0.702} & 1.995 & 0.283 \\
\cmidrule(lr){2-11}
\multirow{3}{*}{\rotatebox[origin=c]{90}{\textbf{GEN}}} 
& SDF-FPC$_{3}$ & 0.527 & 0.926 & 0.200 & 0.692 & 1.914 & 0.246 & 0.699 & 2.029 & 0.287 \\
& SDF-FPC$_{5}$ & 0.530 & 0.918 & 0.198 & 0.693 & 2.003 & 0.252 & 0.662 & \textbf{1.837} & \textbf{0.262} \\
& SDF-FPC$_{7}$ & 0.522 & 0.915 & 0.197 & 0.650 & 1.887 & 0.232 & 0.812 & 2.265 & 0.323 \\
& SDF-ICA$_{3}$ & \textbf{0.514} & 0.899 & 0.194 & 0.647 & 1.829 & 0.233 & 0.730 & 2.068 & 0.294 \\
& SDF-ICA$_{5}$ & 0.537 & 0.909 & 0.194 & 0.637 & 1.934 & 0.233 & \underline{\textbf{0.655}} & 1.849 & \textbf{0.262} \\
& SDF-ICA$_{7}$ & 0.517 & \textbf{0.898} & \textbf{0.193} & \underline{\textbf{0.626}} & \textbf{1.820} & \underline{\textbf{0.224}} & 0.790 & 2.189 & 0.312 \\
\cmidrule(lr){2-11}
\multirow{3}{*}{\rotatebox[origin=c]{90}{\textbf{OG + GEN}}} 
& SDF-FPC$_{3}$ + OD & 0.493 & 0.829 & 0.179 & 0.658 & 1.780 & 0.229 & 0.736 & 2.077 & 0.296 \\
& SDF-FPC$_{5}$ + OD & 0.487 & 0.807 & 0.174 & 0.659 & 1.757 & 0.230 & 0.743 & 2.087 & 0.297 \\
& SDF-FPC$_{7}$ + OD & 0.492 & 0.821 & 0.177 & 0.666 & 1.754 & 0.231 & \textbf{0.706} & \textbf{1.993} & \textbf{0.283} \\
& SDF-ICA$_{3}$ + OD & 0.487 & \underline{\textbf{0.801}} & \underline{\textbf{0.173}} & \textbf{0.640} & 1.790 & 0.228 & 0.734 & 2.074 & 0.295 \\
& SDF-ICA$_{5}$ + OD & \textbf{\underline{0.486}} & 0.804 & 0.174 & 0.649 & 1.780 & 0.230 & 0.750 & 2.110 & 0.301 \\
& SDF-ICA$_{7}$ + OD & 0.490 & 0.810 & 0.175 & 0.642 & \underline{\textbf{1.746}} & \textbf{0.226} & 0.718 & 2.025 & 0.288 \\
\cmidrule[1pt]{2-11}
\end{tabular}
\label{tab:ttm_appendix}
\end{table}

\begin{table}[H]
\caption{\textbf{Average Generation Time Across LLM Backbones.} Average time (in seconds) required to generate synthetic univariate time series for the \texttt{bikesharing} dataset across three targets: \texttt{count}, \texttt{temperature}, and \texttt{humidity}. We report results for two input sequence lengths (250 and 500) and compare three LLM backbones: \texttt{GPT-2}, \texttt{granite-3.0-2b-base}, and \texttt{Phi-3.5-mini-instruct}. All models were evaluated under the same computational constraints (\texttt{-mem 100G -cores 1+1 -gpu a100}) using a single NVIDIA A100 GPU.
For fine-tuning, we use a batch size of 16 for \texttt{granite} and 8 for \texttt{phi}.}
\vspace{0.1cm}
\tiny
\renewcommand{\arraystretch}{1.4}
\rowcolors{2}{gray!20}{white}
\hspace{-0.7cm}
\begin{tabular}{l|rrrrrrrrrrr}
    \toprule
    \textbf{Length} & ICA$_3$ + gpt2 & ICA$_3$ + granite & ICA$_3$ + phi & ICA$_5$ + gpt2 & ICA$_5$ + granite & ICA$_5$ + phi  & ICA$_7$ + gpt2 & ICA$_7$ + granite & ICA$_7$ + phi \\
    \midrule
    \textbf{250} & 22.7 & 112.8 & 132.6 & 18.3 & 118.5 & 113.9 & 19.0 & 119.9 & 126.1 \\
    \textbf{500} & 16.2 & 93.6 & 98.5 & 17.3 & 99.9 & 103.7 & 18.9 & 125.0 & 110.7 \\
    \bottomrule
\end{tabular}
\label{tab:gen_time_llm}
\end{table}

\begin{table}[H]
\caption{\textbf{Ablation study: LLM backbone.} Aggregated similarity-based performance across all datasets 
for different LLMs used in SDF models. We compare \texttt{GPT-2} with two larger and more recent alternatives: \texttt{granite-3.0-2b-base}\tablefootnote{\url{https://huggingface.co/ibm-granite/granite-3.0-2b-base}} (2B parameters) and \texttt{Phi-3.5-mini-instruct}\tablefootnote{\url{https://huggingface.co/microsoft/Phi-3.5-mini-instruct}} (3.8B parameters). For fine-tuning, we use a batch size of 16 for \texttt{granite} and 8 for \texttt{phi}.}
\vspace{0.1cm}
\centering
\tiny
\renewcommand{\arraystretch}{1.4}
\rowcolors{2}{gray!20}{white}
\begin{tabular}{>{\columncolor{white}[-20pt][\tabcolsep]}l lrrrrrrrrr}
    \cmidrule[1pt]{2-11}
    & & \multicolumn{4}{c}{\textbf{Feature-based}} & \multicolumn{3}{c}{\textbf{Distance-based}} & \multicolumn{2}{c}{\textbf{Norm. Avg.}} \\
    \cmidrule(lr){3-6} \cmidrule(lr){7-9} \cmidrule(lr){10-11}
    & & \textbf{MDD} & \textbf{ACD} & \textbf{SD} & \textbf{KD} & \textbf{ED} & \textbf{DTW} & \textbf{SHR} & \textbf{Feat.} & \textbf{Dist.} \\
    \cmidrule{2-11}
\multirow{6}{*}{\rotatebox[origin=c]{90}{\textbf{MULTISAMPLE}}}
    & SDF-FPC$_{3}$ + GPT-2 & 0.255 & 2.166 & 1.323 & 4.299 & 17.749 & 11.921 & 16.537 & 0.964 & 0.747 \\
    & SDF-FPC$_{3}$ + Granite & 0.251 & 1.817 & 1.227 & 4.132 & \underline{16.429} & \textbf{11.659} & 11.565 & 0.757 & \underline{0.245} \\
    & SDF-FPC$_{3}$ + Phi-3 & 0.257 & 1.215 & 1.154 & 3.723 & 16.734 & 11.872 & 12.367 & 0.643 & 0.397 \\
    & SDF-ICA$_{3}$ + GPT-2 & \underline{0.244} & 1.180 & \textbf{0.869} & \textbf{2.384} & 16.669 & 12.373 & 6.870 & \underline{0.101} & 0.361 \\
    & SDF-ICA$_{3}$ + Granite & \textbf{0.241} & \underline{1.069} & \underline{0.961} & \underline{2.524} & 16.953 & 12.744 & \textbf{6.160} & \textbf{0.101} & 0.509 \\
    & SDF-ICA$_{3}$ + Phi-3 & 0.247 & \textbf{0.907} & 1.102 & 3.570 & \textbf{16.069} & \underline{11.847} & \underline{6.499} & 0.382 & \textbf{0.069} \\
\addlinespace[0.5em]
\cmidrule{2-11}
\addlinespace[0.5em]
\multirow{6}{*}{\rotatebox[origin=c]{90}{\textbf{UNIVARIATE}}}
    & SDF-FPC$_{3}$ + GPT-2 & 0.308 & 1.480 & 0.801 & 1.690 & 19.340 & 12.809 & 5.452 & 0.947 & 0.804 \\
    & SDF-FPC$_{3}$ + Granite & \underline{0.305} & \textbf{1.268} & 0.673 & \underline{1.185} & 19.026 & 12.556 & 5.457 & \textbf{0.207} & 0.505 \\
    & SDF-FPC$_{3}$ + Phi-3 & 0.310 & \underline{1.368} & 0.671 & 1.305 & 19.561 & 12.767 & 5.196 & 0.574 & 0.777 \\
    & SDF-ICA$_{3}$ + GPT-2 & 0.306 & 1.396 & \underline{0.671} & 1.382 & \underline{18.802} & \underline{12.435} & \underline{4.856} & 0.496 & \underline{0.161} \\
    & SDF-ICA$_{3}$ + Granite & \textbf{0.304} & 1.370 & 0.679 & \textbf{1.123} & \textbf{18.616} & \textbf{12.365} & \textbf{4.712} & \underline{0.253} & \textbf{0.000} \\
    & SDF-ICA$_{3}$ + Phi-3 & 0.307 & 1.398 & \textbf{0.541} & 1.199 & 19.081 & 12.471 & 5.856 & 0.337 & 0.577 \\
    \cmidrule[1pt]{2-11}
\end{tabular}
\label{tab:ablation_llm_combined}
\end{table}



\begin{table}[H]
\caption{\textbf{Multisample evaluation: similarity metrics reported per dataset.} }
\vspace{0.1cm}
\centering
\tiny
\renewcommand{\arraystretch}{1.4}
\rowcolors{2}{gray!20}{white}
\begin{tabular}{lcccccccc}
    \toprule
    & 
    & \multicolumn{2}{c}{\textbf{SDForger Models}} & \multicolumn{2}{c}{\textbf{VAE Models}} & \multicolumn{2}{c}{\textbf{GAN Models}} & \multicolumn{1}{c}{\textbf{Others}} \\
    \cmidrule(lr){3-4} \cmidrule(lr){5-6} \cmidrule(lr){7-8} \cmidrule(lr){9-9}
    & 
    & \rotatebox{0}{ICA$_3$} & \rotatebox{0}{FPC$_3$}
    & \rotatebox{0}{TimeVAE} & \rotatebox{0}{TimeVQVAE}
    & \rotatebox{0}{RTSGAN} & \rotatebox{0}{SDEGAN} 
    & \rotatebox{0}{LS4} \\
    \midrule
    \begin{tabular}{ll} ECL \end{tabular}
    & \renewcommand{\arraystretch}{0.8} \begin{tabular}{c} MDD \\ ACD \\ SD \\ KD \\ ED \\ DTW \\ SHAP-RE \end{tabular}
    & \renewcommand{\arraystretch}{0.8} \begin{tabular}{c}    \underline{0.154} \\ 0.146 \\ 2.47 \\ 7.342 \\ \underline{10.141} \\ \underline{9.715} \\ \textbf{0.424} \\  \end{tabular}
    & \renewcommand{\arraystretch}{0.8} \begin{tabular}{c}    0.219 \\ 4.492 \\ 2.762 \\ 6.07 \\ 12.543 \\ 10.985 \\ 4.027 \\  \end{tabular}
    & \renewcommand{\arraystretch}{0.8} \begin{tabular}{c}    0.156 \\ \underline{0.082} \\ \underline{0.746} \\ \underline{3.294} \\ 12.887 \\ 12.025 \\ \underline{1.922} \\  \end{tabular}
    & \renewcommand{\arraystretch}{0.8} \begin{tabular}{c}    0.292 \\ 7.862 \\ 2.803 \\ 6.895 \\ \textbf{9.978} \\ \textbf{7.97} \\ 2.21 \\  \end{tabular}
    & \renewcommand{\arraystretch}{0.8} \begin{tabular}{c}    \textbf{0.145} \\ \textbf{0.051} \\ \textbf{0.114} \\ \textbf{1.001} \\ 13.461 \\ 12.486 \\ 4.123 \\  \end{tabular}
    & \renewcommand{\arraystretch}{0.8} \begin{tabular}{c}    0.193 \\ 0.174 \\ 2.91 \\ 8.673 \\ 25.951 \\ 25.431 \\ 7.048 \\  \end{tabular}
    & \renewcommand{\arraystretch}{0.8} \begin{tabular}{c}    0.296 \\ 8.365 \\ 2.983 \\ 10.074 \\ 43.715 \\ 40.142 \\ 105.287 \\  \end{tabular}
    \\
    \begin{tabular}{ll} NN5 \end{tabular}
    & \renewcommand{\arraystretch}{0.8} \begin{tabular}{c} MDD \\ ACD \\ SD \\ KD \\ ED \\ DTW \\ SHAP-RE \end{tabular}
    & \renewcommand{\arraystretch}{0.8} \begin{tabular}{c}    0.248 \\ \underline{1.489} \\ 0.307 \\ 1.43 \\ 19.308 \\ 12.837 \\ 9.482 \\  \end{tabular}
    & \renewcommand{\arraystretch}{0.8} \begin{tabular}{c}    0.248 \\ 3.235 \\ 0.428 \\ 4.249 \\ 20.576 \\ 12.617 \\ 40.975 \\  \end{tabular}
    & \renewcommand{\arraystretch}{0.8} \begin{tabular}{c}    \textbf{0.243} \\ \textbf{0.221} \\ \underline{0.126} \\ \textbf{0.151} \\ 20.514 \\ 11.223 \\ 20.993 \\  \end{tabular}
    & \renewcommand{\arraystretch}{0.8} \begin{tabular}{c}    0.371 \\ 4.964 \\ 0.259 \\ 1.159 \\ \textbf{15.019} \\ \underline{10.914} \\ \textbf{2.072} \\  \end{tabular}
    & \renewcommand{\arraystretch}{0.8} \begin{tabular}{c}    0.383 \\ 4.646 \\ \textbf{0.092} \\ \underline{0.348} \\ \underline{16.201} \\ \textbf{9.684} \\ \underline{8.254} \\  \end{tabular}
    & \renewcommand{\arraystretch}{0.8} \begin{tabular}{c}    \underline{0.246} \\ 2.73 \\ 0.422 \\ 0.512 \\ 43.433 \\ 36.712 \\ 83.918 \\  \end{tabular}
    & \renewcommand{\arraystretch}{0.8} \begin{tabular}{c}    0.262 \\ 5.677 \\ 0.287 \\ 0.96 \\ 38.822 \\ 24.415 \\ 207.419 \\  \end{tabular}
    
    \\
    \begin{tabular}{ll} Tourism \end{tabular}
    & \renewcommand{\arraystretch}{0.8} \begin{tabular}{c} MDD \\ ACD \\ SD \\ KD \\ ED \\ DTW \\ SHAP-RE \end{tabular}
    & \renewcommand{\arraystretch}{0.8} \begin{tabular}{c}    0.189 \\ 0.22 \\ 1.321 \\ \underline{4.477} \\ \underline{11.216} \\ \underline{10.291} \\ \textbf{0.547} \\  \end{tabular}
    & \renewcommand{\arraystretch}{0.8} \begin{tabular}{c}    0.24 \\ 3.29 \\ 2.613 \\ 8.297 \\ 14.039 \\ 11.837 \\ 4.28 \\  \end{tabular}
    & \renewcommand{\arraystretch}{0.8} \begin{tabular}{c}    \underline{0.172} \\ \textbf{0.206} \\ \underline{0.854} \\ \textbf{4.251} \\ 13.895 \\ 12.409 \\ \underline{1.785} \\  \end{tabular}
    & \renewcommand{\arraystretch}{0.8} \begin{tabular}{c}    0.339 \\ 7.807 \\ 2.896 \\ 7.877 \\ \textbf{10.516} \\ \textbf{8.192} \\ 2.135 \\  \end{tabular}
    & \renewcommand{\arraystretch}{0.8} \begin{tabular}{c}    \textbf{0.121} \\ 0.272 \\ \textbf{0.681} \\ 4.852 \\ 15.405 \\ 14.741 \\ 2.833 \\  \end{tabular}
    & \renewcommand{\arraystretch}{0.8} \begin{tabular}{c}    0.208 \\ \underline{0.215} \\ 2.988 \\ 9.616 \\ 25.399 \\ 25.098 \\ 6.409 \\  \end{tabular}
    & \renewcommand{\arraystretch}{0.8} \begin{tabular}{c}    0.282 \\ 8.228 \\ 2.806 \\ 10.907 \\ 39.897 \\ 35.521 \\ 100.231 \\  \end{tabular}
    
    \\
    \begin{tabular}{ll} Traffic \end{tabular}
    & \renewcommand{\arraystretch}{0.8} \begin{tabular}{c} MDD \\ ACD \\ SD \\ KD \\ ED \\ DTW \\ SHAP-RE \end{tabular}
    & \renewcommand{\arraystretch}{0.8} \begin{tabular}{c}    0.251 \\ 1.443 \\ 1.433 \\ 3.177 \\ 16.532 \\ 10.917 \\ 6.568 \\  \end{tabular}
    & \renewcommand{\arraystretch}{0.8} \begin{tabular}{c}    \underline{0.234} \\ 1.368 \\ 1.507 \\ 1.937 \\ 18.429 \\ 10.5 \\ 9.188 \\  \end{tabular}
    & \renewcommand{\arraystretch}{0.8} \begin{tabular}{c}    \underline{0.234} \\ \textbf{0.097} \\ \textbf{0.353} \\ \textbf{1.263} \\ 20.748 \\ 12.343 \\ 14.948 \\  \end{tabular}
    & \renewcommand{\arraystretch}{0.8} \begin{tabular}{c}    0.359 \\ 3.767 \\ \underline{1.377} \\ \underline{1.503} \\ \textbf{14.169} \\ \underline{10.039} \\ \textbf{1.995} \\  \end{tabular}
    & \renewcommand{\arraystretch}{0.8} \begin{tabular}{c}    0.314 \\ \underline{0.886} \\ 1.384 \\ 2.642 \\ \underline{15.552} \\ \textbf{9.164} \\ \underline{5.101} \\  \end{tabular}
    & \renewcommand{\arraystretch}{0.8} \begin{tabular}{c}    \textbf{0.222} \\ 3.749 \\ 1.598 \\ 3.113 \\ 35.522 \\ 31.69 \\ 51.49 \\  \end{tabular}
    & \renewcommand{\arraystretch}{0.8} \begin{tabular}{c}    0.242 \\ 5.229 \\ 1.403 \\ 4.727 \\ 55.908 \\ 37.028 \\ 205.07 \\  \end{tabular}
    
    \\
    \begin{tabular}{ll} Weather \\ (Maxtemp) \end{tabular}
    & \renewcommand{\arraystretch}{0.8} \begin{tabular}{c} MDD \\ ACD \\ SD \\ KD \\ ED \\ DTW \\ SHAP-RE \end{tabular}
    & \renewcommand{\arraystretch}{0.8} \begin{tabular}{c}    0.292 \\ 1.097 \\ 0.131 \\ \textbf{0.419} \\ 25.51 \\ 20.532 \\ 6.638 \\  \end{tabular}
    & \renewcommand{\arraystretch}{0.8} \begin{tabular}{c}    0.293 \\ 2.221 \\ \underline{0.017} \\ 2.561 \\ 21.661 \\ 15.354 \\ 24.439 \\  \end{tabular}
    & \renewcommand{\arraystretch}{0.8} \begin{tabular}{c}    \textbf{0.282} \\ \textbf{0.533} \\ 0.435 \\ 0.591 \\ \underline{19.665} \\ \underline{13.238} \\ 9.079 \\  \end{tabular}
    & \renewcommand{\arraystretch}{0.8} \begin{tabular}{c}    0.447 \\ 6.717 \\ \textbf{0.005} \\ 2.188 \\ \textbf{15.113} \\ \textbf{11.219} \\ \textbf{2.098} \\  \end{tabular}
    & \renewcommand{\arraystretch}{0.8} \begin{tabular}{c}    0.303 \\ 1.15 \\ 0.37 \\ 0.584 \\ 19.77 \\ 15.884 \\ \underline{3.931} \\  \end{tabular}
    & \renewcommand{\arraystretch}{0.8} \begin{tabular}{c}    \underline{0.286} \\ \underline{0.705} \\ 0.237 \\ \underline{0.527} \\ 43.035 \\ 41.331 \\ 34.196 \\  \end{tabular}
    & \renewcommand{\arraystretch}{0.8} \begin{tabular}{c}    0.296 \\ 7.597 \\ 0.142 \\ 0.976 \\ 46.189 \\ 34.55 \\ 152.274 \\  \end{tabular}
    
    \\
    \begin{tabular}{ll} Weather \\ (Mintemp) \end{tabular}
    & \renewcommand{\arraystretch}{0.8} \begin{tabular}{c} MDD \\ ACD \\ SD \\ KD \\ ED \\ DTW \\ SHAP-RE \end{tabular}
    & \renewcommand{\arraystretch}{0.8} \begin{tabular}{c}    0.271 \\ 1.705 \\ 0.185 \\ 1.065 \\ 17.199 \\ 13.148 \\ 6.217 \\  \end{tabular}
    & \renewcommand{\arraystretch}{0.8} \begin{tabular}{c}    0.27 \\ \underline{1.053} \\ 0.216 \\ 1.789 \\ 18.533 \\ 11.255 \\ 14.613 \\  \end{tabular}
    & \renewcommand{\arraystretch}{0.8} \begin{tabular}{c}    \textbf{0.264} \\ \textbf{0.278} \\ 0.304 \\ \underline{0.757} \\ 19.575 \\ \textbf{10.48} \\ 15.605 \\  \end{tabular}
    & \renewcommand{\arraystretch}{0.8} \begin{tabular}{c}    0.41 \\ 5.937 \\ \textbf{0.05} \\ 1.936 \\ \textbf{15.177} \\ \underline{10.675} \\ \textbf{1.956} \\  \end{tabular}
    & \renewcommand{\arraystretch}{0.8} \begin{tabular}{c}    0.365 \\ 1.508 \\ 0.398 \\ 0.985 \\ \underline{15.575} \\ 10.837 \\ \underline{2.587} \\  \end{tabular}
    & \renewcommand{\arraystretch}{0.8} \begin{tabular}{c}    \underline{0.266} \\ 1.67 \\ 0.249 \\ \textbf{0.213} \\ 44.094 \\ 41.651 \\ 55.846 \\  \end{tabular}
    & \renewcommand{\arraystretch}{0.8} \begin{tabular}{c}    0.275 \\ 6.737 \\ \underline{0.145} \\ 1.174 \\ 46.266 \\ 32.415 \\ 208.289 \\  \end{tabular}
    
    \\
    \begin{tabular}{ll} Weather \\ (Rain) \end{tabular}
    & \renewcommand{\arraystretch}{0.8} \begin{tabular}{c} MDD \\ ACD \\ SD \\ KD \\ ED \\ DTW \\ SHAP-RE \end{tabular}
    & \renewcommand{\arraystretch}{0.8} \begin{tabular}{c}    0.233 \\ 1.642 \\ \textbf{0.599} \\ \textbf{1.146} \\ \underline{14.461} \\ 10.935 \\ 10.955 \\  \end{tabular}
    & \renewcommand{\arraystretch}{0.8} \begin{tabular}{c}    0.222 \\ \underline{0.321} \\ 2.474 \\ 6.636 \\ 16.009 \\ 11.493 \\ \underline{9.555} \\  \end{tabular}
    & \renewcommand{\arraystretch}{0.8} \begin{tabular}{c}    \textbf{0.175} \\ \textbf{0.309} \\ \underline{1.065} \\ \underline{3.238} \\ 16.979 \\ 11.35 \\ 25.464 \\  \end{tabular}
    & \renewcommand{\arraystretch}{0.8} \begin{tabular}{c}    0.291 \\ 2.491 \\ 2.683 \\ 7.94 \\ \textbf{13.631} \\ \textbf{10.718} \\ \textbf{1.883} \\  \end{tabular}
    & \renewcommand{\arraystretch}{0.8} \begin{tabular}{c}    0.259 \\ 2.735 \\ 1.641 \\ 7.447 \\ 15.033 \\ \underline{10.803} \\ 9.595 \\  \end{tabular}
    & \renewcommand{\arraystretch}{0.8} \begin{tabular}{c}    \underline{0.199} \\ 5.549 \\ 2.729 \\ 9.639 \\ 31.412 \\ 28.891 \\ 57.118 \\  \end{tabular}
    & \renewcommand{\arraystretch}{0.8} \begin{tabular}{c}    0.27 \\ 4.696 \\ 2.135 \\ 9.724 \\ 50.966 \\ 33.017 \\ 224.024 \\  \end{tabular}    
    \\
    \begin{tabular}{ll} Weather \\ (Solar) \end{tabular}
    & \renewcommand{\arraystretch}{0.8} \begin{tabular}{c} MDD \\ ACD \\ SD \\ KD \\ ED \\ DTW \\ SHAP-RE \end{tabular}
    & \renewcommand{\arraystretch}{0.8} \begin{tabular}{c}    0.314 \\ 1.695 \\ 0.508 \\ \textbf{0.013} \\ 18.989 \\ \underline{10.61} \\ 14.132 \\  \end{tabular}
    & \renewcommand{\arraystretch}{0.8} \begin{tabular}{c}    0.31 \\ \underline{1.351} \\ 0.564 \\ 2.854 \\ 20.201 \\ 11.322 \\ 25.218 \\  \end{tabular}
    & \renewcommand{\arraystretch}{0.8} \begin{tabular}{c}    \underline{0.29} \\ \textbf{0.349} \\ 0.172 \\ \underline{0.029} \\ 20.061 \\ \textbf{9.932} \\ 22.375 \\  \end{tabular}
    & \renewcommand{\arraystretch}{0.8} \begin{tabular}{c}    0.459 \\ 4.18 \\ 0.539 \\ 1.618 \\ \textbf{15.69} \\ 11.607 \\ \textbf{1.892} \\  \end{tabular}
    & \renewcommand{\arraystretch}{0.8} \begin{tabular}{c}    0.342 \\ 2.902 \\ 0.216 \\ 0.544 \\ \underline{17.677} \\ 11.275 \\ \underline{8.623} \\  \end{tabular}
    & \renewcommand{\arraystretch}{0.8} \begin{tabular}{c}    0.297 \\ 1.996 \\ \underline{0.101} \\ 0.433 \\ 48.549 \\ 36.321 \\ 117.395 \\  \end{tabular}
    & \renewcommand{\arraystretch}{0.8} \begin{tabular}{c}    \textbf{0.284} \\ 2.667 \\ \textbf{0.044} \\ 0.27 \\ 33.346 \\ 17.363 \\ 80.632 \\  \end{tabular}
    \\
    \bottomrule
\end{tabular}
\label{tab:multisample_per_data}
\end{table}

\begin{table}[H]
\caption{\textbf{Univariate evaluation: similarity metrics reported for Energy datasets.}}
\vspace{0.1cm}
\centering
\tiny
\renewcommand{\arraystretch}{1.4}
\rowcolors{2}{gray!20}{white}
\begin{tabular}{lccccccccc}
    \toprule
    & 
    & \multicolumn{2}{c}{\textbf{SDForger Models}} & \multicolumn{2}{c}{\textbf{VAE Models}} & \multicolumn{2}{c}{\textbf{GAN Models}} & \multicolumn{1}{c}{\textbf{Others}} \\
    \cmidrule(lr){3-4} \cmidrule(lr){5-6} \cmidrule(lr){7-8} \cmidrule(lr){9-9}
    & 
    & \rotatebox{0}{ICA$_3$} & \rotatebox{0}{FPC$_3$}
    & \rotatebox{0}{TimeVAE} & \rotatebox{0}{TimeVQVAE}
    & \rotatebox{0}{RTSGAN} & \rotatebox{0}{SDEGAN} 
    & \rotatebox{0}{LS4} \\
    \midrule
    \begin{tabular}{ll} Appliances \end{tabular}
    & \renewcommand{\arraystretch}{0.8} \begin{tabular}{c} MDD \\ ACD \\ SD \\ KD \\ ED \\ DTW \\ SHAP-RE \end{tabular}
    & \renewcommand{\arraystretch}{0.8} \begin{tabular}{c}    0.318 \\ \underline{1.825} \\ 2.068 \\ 3.249 \\ 19.493 \\ 11.597 \\ 9.588 \\  \end{tabular}
    & \renewcommand{\arraystretch}{0.8} \begin{tabular}{c}    0.315 \\ \textbf{1.714} \\ 2.093 \\ 3.816 \\ 19.13 \\ 11.73 \\ 9.072 \\  \end{tabular}
    & \renewcommand{\arraystretch}{0.8} \begin{tabular}{c}    0.303 \\ 3.137 \\ \textbf{1.008} \\ 2.891 \\ 20.459 \\ \textbf{10.03} \\ 27.262 \\  \end{tabular}
    & \renewcommand{\arraystretch}{0.8} \begin{tabular}{c}    0.405 \\ \textbf{1.714} \\ 2.07 \\ \underline{2.586} \\ \textbf{15.916} \\ 12.322 \\ \textbf{1.952} \\  \end{tabular}
    & \renewcommand{\arraystretch}{0.8} \begin{tabular}{c}    0.414 \\ 2.404 \\ 2.152 \\ \textbf{2.559} \\ \underline{17.162} \\ \underline{11.301} \\ \underline{6.367} \\  \end{tabular}
    & \renewcommand{\arraystretch}{0.8} \begin{tabular}{c}    \textbf{0.241} \\ 7.208 \\ 2.269 \\ 4.196 \\ 29.562 \\ 27.191 \\ 71.477 \\  \end{tabular}
    & \renewcommand{\arraystretch}{0.8} \begin{tabular}{c}    \underline{0.251} \\ 4.301 \\ \underline{2.028} \\ 5.874 \\ 47.81 \\ 16.564 \\ 162.749 \\  \end{tabular}
    \\
    \begin{tabular}{ll} Australian Elec \\ (T000000) \end{tabular}
    & \renewcommand{\arraystretch}{0.8} \begin{tabular}{c} MDD \\ ACD \\ SD \\ KD \\ ED \\ DTW \\ SHAP-RE \end{tabular}
    & \renewcommand{\arraystretch}{0.8} \begin{tabular}{c}    0.315 \\ \underline{1.069} \\ 0.165 \\ \textbf{0.161} \\ \underline{19.017} \\ \underline{10.652} \\ 3.411 \\  \end{tabular}
    & \renewcommand{\arraystretch}{0.8} \begin{tabular}{c}    0.312 \\ \textbf{0.392} \\ 0.103 \\ 0.452 \\ 22.663 \\ 13.655 \\ \underline{2.719} \\  \end{tabular}
    & \renewcommand{\arraystretch}{0.8} \begin{tabular}{c}    \underline{0.292} \\ 1.431 \\ 0.65 \\ 1.296 \\ 19.656 \\ 10.899 \\ 7.925 \\  \end{tabular}
    & \renewcommand{\arraystretch}{0.8} \begin{tabular}{c}    0.457 \\ 5.641 \\ \textbf{0.071} \\ 2.12 \\ \textbf{15.657} \\ \textbf{10.583} \\ \textbf{2.441} \\  \end{tabular}
    & \renewcommand{\arraystretch}{0.8} \begin{tabular}{c}    0.376 \\ 1.926 \\ 0.3 \\ 1.829 \\ 22.511 \\ 14.942 \\ 4.007 \\  \end{tabular}
    & \renewcommand{\arraystretch}{0.8} \begin{tabular}{c}    \textbf{0.281} \\ 2.976 \\ \underline{0.091} \\ \underline{0.325} \\ 42.67 \\ 40.842 \\ 23.325 \\  \end{tabular}
    & \renewcommand{\arraystretch}{0.8} \begin{tabular}{c}    0.323 \\ 7.129 \\ 0.423 \\ 0.765 \\ 38.681 \\ 26.104 \\ 34.777 \\  \end{tabular}
    \\
    \begin{tabular}{ll} Australian Elec \\ (T000001) \end{tabular}
    & \renewcommand{\arraystretch}{0.8} \begin{tabular}{c} MDD \\ ACD \\ SD \\ KD \\ ED \\ DTW \\ SHAP-RE \end{tabular}
    & \renewcommand{\arraystretch}{0.8} \begin{tabular}{c}    0.287 \\ 1.46 \\ 0.274 \\ 1.8 \\ \underline{18.368} \\ 12.248 \\ \textbf{2.211} \\  \end{tabular}
    & \renewcommand{\arraystretch}{0.8} \begin{tabular}{c}    0.281 \\ \underline{0.457} \\ 0.307 \\ 1.677 \\ 19.769 \\ 12.842 \\ 2.91 \\  \end{tabular}
    & \renewcommand{\arraystretch}{0.8} \begin{tabular}{c}    \underline{0.277} \\ \textbf{0.265} \\ 0.11 \\ 0.503 \\ 19.51 \\ \textbf{10.312} \\ 6.061 \\  \end{tabular}
    & \renewcommand{\arraystretch}{0.8} \begin{tabular}{c}    0.436 \\ 5.579 \\ 0.137 \\ 1.99 \\ \textbf{15.717} \\ \underline{10.648} \\ \underline{2.587} \\  \end{tabular}
    & \renewcommand{\arraystretch}{0.8} \begin{tabular}{c}    0.421 \\ 4.796 \\ 0.589 \\ \textbf{0.264} \\ 20.145 \\ 13.322 \\ 3.291 \\  \end{tabular}
    & \renewcommand{\arraystretch}{0.8} \begin{tabular}{c}    \textbf{0.273} \\ 2.518 \\ \textbf{0.031} \\ \underline{0.273} \\ 44.034 \\ 41.201 \\ 26.765 \\  \end{tabular}
    & \renewcommand{\arraystretch}{0.8} \begin{tabular}{c}    0.293 \\ 6.908 \\ \underline{0.032} \\ 0.899 \\ 37.357 \\ 25.029 \\ 30.768 \\  \end{tabular}
    \\
    \begin{tabular}{ll} Australian Elec \\ (T000002) \end{tabular}
    & \renewcommand{\arraystretch}{0.8} \begin{tabular}{c} MDD \\ ACD \\ SD \\ KD \\ ED \\ DTW \\ SHAP-RE \end{tabular}
    & \renewcommand{\arraystretch}{0.8} \begin{tabular}{c}    0.313 \\ 1.196 \\ 0.42 \\ \textbf{0.179} \\ 20.624 \\ 13.17 \\ 2.683 \\  \end{tabular}
    & \renewcommand{\arraystretch}{0.8} \begin{tabular}{c}    0.313 \\ \underline{0.569} \\ 0.718 \\ 0.69 \\ 21.174 \\ 13.104 \\ 3.04 \\  \end{tabular}
    & \renewcommand{\arraystretch}{0.8} \begin{tabular}{c}    \textbf{0.292} \\ \textbf{0.237} \\ \underline{0.322} \\ 0.539 \\ \underline{19.537} \\ \textbf{10.626} \\ 4.984 \\  \end{tabular}
    & \renewcommand{\arraystretch}{0.8} \begin{tabular}{c}    0.486 \\ 5.675 \\ 0.598 \\ 2.111 \\ \textbf{15.889} \\ \underline{10.748} \\ \underline{2.38} \\  \end{tabular}
    & \renewcommand{\arraystretch}{0.8} \begin{tabular}{c}    0.334 \\ 2.541 \\ 0.605 \\ 2.386 \\ 19.896 \\ 14.559 \\ \textbf{1.572} \\  \end{tabular}
    & \renewcommand{\arraystretch}{0.8} \begin{tabular}{c}    \underline{0.297} \\ 2.559 \\ 0.387 \\ \underline{0.389} \\ 45.934 \\ 44.425 \\ 23.29 \\  \end{tabular}
    & \renewcommand{\arraystretch}{0.8} \begin{tabular}{c}    0.315 \\ 7.209 \\ \textbf{0.229} \\ 0.804 \\ 37.731 \\ 26.567 \\ 28.057 \\  \end{tabular}
    \\
    \begin{tabular}{ll} Australian Elec \\ (T000003) \end{tabular}
    & \renewcommand{\arraystretch}{0.8} \begin{tabular}{c} MDD \\ ACD \\ SD \\ KD \\ ED \\ DTW \\ SHAP-RE \end{tabular}
    & \renewcommand{\arraystretch}{0.8} \begin{tabular}{c}    0.283 \\ 1.549 \\ 0.399 \\ \textbf{0.057} \\ \underline{17.406} \\ 11.621 \\ \textbf{1.888} \\  \end{tabular}
    & \renewcommand{\arraystretch}{0.8} \begin{tabular}{c}    0.289 \\ \underline{1.35} \\ 0.389 \\ 0.465 \\ 19.377 \\ 14.456 \\ \underline{1.929} \\  \end{tabular}
    & \renewcommand{\arraystretch}{0.8} \begin{tabular}{c}    \underline{0.279} \\ \textbf{0.455} \\ \underline{0.135} \\ 0.226 \\ 20.146 \\ \underline{11.466} \\ 4.942 \\  \end{tabular}
    & \renewcommand{\arraystretch}{0.8} \begin{tabular}{c}    0.444 \\ 5.174 \\ 0.221 \\ 1.596 \\ \textbf{15.975} \\ \textbf{11.025} \\ 2.522 \\  \end{tabular}
    & \renewcommand{\arraystretch}{0.8} \begin{tabular}{c}    0.41 \\ 4.187 \\ 0.822 \\ 0.524 \\ 18.751 \\ 11.502 \\ 3.095 \\  \end{tabular}
    & \renewcommand{\arraystretch}{0.8} \begin{tabular}{c}    \textbf{0.261} \\ 2.791 \\ 0.365 \\ \underline{0.127} \\ 40.657 \\ 37.719 \\ 17.34 \\  \end{tabular}
    & \renewcommand{\arraystretch}{0.8} \begin{tabular}{c}    0.286 \\ 6.412 \\ \textbf{0.034} \\ 1.343 \\ 40.4 \\ 28.914 \\ 26.566 \\  \end{tabular}
    \\
    \begin{tabular}{ll} Australian Elec \\ (T000004) \end{tabular}
    & \renewcommand{\arraystretch}{0.8} \begin{tabular}{c} MDD \\ ACD \\ SD \\ KD \\ ED \\ DTW \\ SHAP-RE \end{tabular}
    & \renewcommand{\arraystretch}{0.8} \begin{tabular}{c}    0.293 \\ \underline{1.68} \\ 0.088 \\ \textbf{0.199} \\ 19.85 \\ 11.8 \\ \underline{4.11} \\  \end{tabular}
    & \renewcommand{\arraystretch}{0.8} \begin{tabular}{c}    0.287 \\ \textbf{0.887} \\ \underline{0.062} \\ 0.664 \\ 19.864 \\ 12.582 \\ 4.468 \\  \end{tabular}
    & \renewcommand{\arraystretch}{0.8} \begin{tabular}{c}    \underline{0.28} \\ 1.735 \\ 0.666 \\ 1.679 \\ 20.51 \\ \textbf{10.921} \\ 11.67 \\  \end{tabular}
    & \renewcommand{\arraystretch}{0.8} \begin{tabular}{c}    0.458 \\ 4.565 \\ 0.136 \\ 2.199 \\ \textbf{15.88} \\ \underline{10.933} \\ \textbf{2.523} \\  \end{tabular}
    & \renewcommand{\arraystretch}{0.8} \begin{tabular}{c}    0.337 \\ 2.658 \\ 0.173 \\ 0.691 \\ \underline{19.327} \\ 12.972 \\ 4.131 \\  \end{tabular}
    & \renewcommand{\arraystretch}{0.8} \begin{tabular}{c}    \textbf{0.277} \\ 3.209 \\ \textbf{0.057} \\ \underline{0.527} \\ 44.238 \\ 40.349 \\ 39.504 \\  \end{tabular}
    & \renewcommand{\arraystretch}{0.8} \begin{tabular}{c}    0.302 \\ 6.726 \\ 0.365 \\ 0.706 \\ 41.623 \\ 28.13 \\ 54.245 \\  \end{tabular}
    \\
    \begin{tabular}{ll} ETTH1 \\ (HUFL) \end{tabular}
    & \renewcommand{\arraystretch}{0.8} \begin{tabular}{c} MDD \\ ACD \\ SD \\ KD \\ ED \\ DTW \\ SHAP-RE \end{tabular}
    & \renewcommand{\arraystretch}{0.8} \begin{tabular}{c}    0.326 \\ \textbf{1.249} \\ 0.355 \\ \underline{0.745} \\ 19.324 \\ \textbf{9.475} \\ 9.878 \\  \end{tabular}
    & \renewcommand{\arraystretch}{0.8} \begin{tabular}{c}    0.32 \\ \underline{1.376} \\ 0.45 \\ 0.845 \\ 20.103 \\ \underline{9.579} \\ 13.289 \\  \end{tabular}
    & \renewcommand{\arraystretch}{0.8} \begin{tabular}{c}    \textbf{0.29} \\ 3.618 \\ 0.956 \\ 1.5 \\ 21.804 \\ 10.69 \\ 28.806 \\  \end{tabular}
    & \renewcommand{\arraystretch}{0.8} \begin{tabular}{c}    0.476 \\ 2.678 \\ 0.371 \\ 2.067 \\ \textbf{16.021} \\ 11.35 \\ \textbf{1.94} \\  \end{tabular}
    & \renewcommand{\arraystretch}{0.8} \begin{tabular}{c}    0.381 \\ 4.233 \\ \underline{0.153} \\ 0.869 \\ \underline{17.323} \\ 10.616 \\ \underline{7.684} \\  \end{tabular}
    & \renewcommand{\arraystretch}{0.8} \begin{tabular}{c}    \underline{0.291} \\ 4.542 \\ \textbf{0.142} \\ 1.0 \\ 45.316 \\ 38.346 \\ 129.161 \\  \end{tabular}
    & \renewcommand{\arraystretch}{0.8} \begin{tabular}{c}    0.326 \\ 4.778 \\ 0.751 \\ \textbf{0.276} \\ 25.175 \\ 9.745 \\ 48.028 \\  \end{tabular}
    \\
    \begin{tabular}{ll} ETTH1 \\ (OT) \end{tabular}
    & \renewcommand{\arraystretch}{0.8} \begin{tabular}{c} MDD \\ ACD \\ SD \\ KD \\ ED \\ DTW \\ SHAP-RE \end{tabular}
    & \renewcommand{\arraystretch}{0.8} \begin{tabular}{c}    \underline{0.235} \\ 2.276 \\ \textbf{0.104} \\ 0.892 \\ 20.481 \\ 12.525 \\ \underline{3.368} \\  \end{tabular}
    & \renewcommand{\arraystretch}{0.8} \begin{tabular}{c}    0.24 \\ \textbf{1.163} \\ 0.179 \\ \textbf{0.097} \\ \underline{18.292} \\ 12.195 \\ 3.444 \\  \end{tabular}
    & \renewcommand{\arraystretch}{0.8} \begin{tabular}{c}    \textbf{0.225} \\ \underline{1.923} \\ 1.006 \\ 0.651 \\ 22.676 \\ 12.614 \\ 15.306 \\  \end{tabular}
    & \renewcommand{\arraystretch}{0.8} \begin{tabular}{c}    0.354 \\ 4.969 \\ 0.332 \\ 1.089 \\ \textbf{15.583} \\ \textbf{10.499} \\ \textbf{2.174} \\  \end{tabular}
    & \renewcommand{\arraystretch}{0.8} \begin{tabular}{c}    0.268 \\ 2.67 \\ 0.76 \\ 0.936 \\ 20.463 \\ \underline{11.718} \\ 5.613 \\  \end{tabular}
    & \renewcommand{\arraystretch}{0.8} \begin{tabular}{c}    0.254 \\ 3.337 \\ 0.176 \\ \underline{0.645} \\ 51.747 \\ 49.511 \\ 52.871 \\  \end{tabular}
    & \renewcommand{\arraystretch}{0.8} \begin{tabular}{c}    0.272 \\ 6.463 \\ \underline{0.134} \\ 1.885 \\ 42.246 \\ 25.674 \\ 52.186 \\  \end{tabular}
    \\
    \bottomrule
\end{tabular}
\label{tab:univariate_per_data_energy}
\end{table}

\begin{table}[H]
\caption{\textbf{Univariate evaluation: similarity metrics reported for Transport datasets.}}
\vspace{0.1cm}
\centering
\tiny
\renewcommand{\arraystretch}{1.4}
\rowcolors{2}{gray!20}{white}
\begin{tabular}{lccccccccc}
    \toprule
    & 
    & \multicolumn{2}{c}{\textbf{SDForger Models}} & \multicolumn{2}{c}{\textbf{VAE Models}} & \multicolumn{2}{c}{\textbf{GAN Models}} & \multicolumn{1}{c}{\textbf{Others}} \\
    \cmidrule(lr){3-4} \cmidrule(lr){5-6} \cmidrule(lr){7-8} \cmidrule(lr){9-9}
    & 
    & \rotatebox{0}{ICA$_3$} & \rotatebox{0}{FPC$_3$}
    & \rotatebox{0}{TimeVAE} & \rotatebox{0}{TimeVQVAE}
    & \rotatebox{0}{RTSGAN} & \rotatebox{0}{SDEGAN} 
    & \rotatebox{0}{LS4} \\
    \midrule
    \begin{tabular}{ll} Bikesharing \\ (Count) \end{tabular}
    & \renewcommand{\arraystretch}{0.8} \begin{tabular}{c} MDD \\ ACD \\ SD \\ KD \\ ED \\ DTW \\ SHAP-RE \end{tabular}
    & \renewcommand{\arraystretch}{0.8} \begin{tabular}{c}    0.326 \\ \underline{0.856} \\ 0.101 \\ 0.571 \\ 21.389 \\ 10.305 \\ \underline{13.665} \\  \end{tabular}
    & \renewcommand{\arraystretch}{0.8} \begin{tabular}{c}    0.336 \\ \textbf{0.691} \\ 0.12 \\ 0.258 \\ \underline{19.406} \\ \textbf{10.036} \\ 13.77 \\  \end{tabular}
    & \renewcommand{\arraystretch}{0.8} \begin{tabular}{c}    \underline{0.295} \\ 2.493 \\ 0.202 \\ \underline{0.125} \\ 21.314 \\ \underline{10.163} \\ 38.651 \\  \end{tabular}
    & \renewcommand{\arraystretch}{0.8} \begin{tabular}{c}    0.492 \\ 1.712 \\ \underline{0.042} \\ 1.447 \\ \textbf{16.172} \\ 12.804 \\ \textbf{2.012} \\  \end{tabular}
    & \renewcommand{\arraystretch}{0.8} \begin{tabular}{c}    0.444 \\ 2.451 \\ 0.251 \\ \textbf{0.091} \\ 19.806 \\ 10.225 \\ 19.352 \\  \end{tabular}
    & \renewcommand{\arraystretch}{0.8} \begin{tabular}{c}    \textbf{0.29} \\ 4.306 \\ 0.61 \\ 0.218 \\ 44.217 \\ 31.743 \\ 247.88 \\  \end{tabular}
    & \renewcommand{\arraystretch}{0.8} \begin{tabular}{c}    0.332 \\ 1.972 \\ \textbf{0.035} \\ 0.387 \\ 21.848 \\ 10.347 \\ 64.166 \\  \end{tabular}
    \\
    \begin{tabular}{ll} Bikesharing \\ (Humidity) \end{tabular}
    & \renewcommand{\arraystretch}{0.8} \begin{tabular}{c} MDD \\ ACD \\ SD \\ KD \\ ED \\ DTW \\ SHAP-RE \end{tabular}
    & \renewcommand{\arraystretch}{0.8} \begin{tabular}{c}    0.318 \\ \textbf{1.653} \\ 0.116 \\ 0.521 \\ 19.761 \\ \underline{10.699} \\ \underline{4.249} \\  \end{tabular}
    & \renewcommand{\arraystretch}{0.8} \begin{tabular}{c}    0.322 \\ \underline{1.94} \\ \underline{0.113} \\ 0.806 \\ \underline{18.098} \\ 10.847 \\ 4.63 \\  \end{tabular}
    & \renewcommand{\arraystretch}{0.8} \begin{tabular}{c}    0.301 \\ 4.68 \\ 0.569 \\ 0.671 \\ 25.8 \\ 13.416 \\ 43.651 \\  \end{tabular}
    & \renewcommand{\arraystretch}{0.8} \begin{tabular}{c}    0.432 \\ 2.715 \\ \textbf{0.076} \\ 1.831 \\ \textbf{16.097} \\ 10.907 \\ \textbf{2.154} \\  \end{tabular}
    & \renewcommand{\arraystretch}{0.8} \begin{tabular}{c}    0.365 \\ 3.323 \\ 0.367 \\ \underline{0.511} \\ 18.478 \\ \textbf{10.25} \\ 7.005 \\  \end{tabular}
    & \renewcommand{\arraystretch}{0.8} \begin{tabular}{c}    \textbf{0.266} \\ 4.948 \\ 0.425 \\ \textbf{0.404} \\ 44.641 \\ 36.872 \\ 88.647 \\  \end{tabular}
    & \renewcommand{\arraystretch}{0.8} \begin{tabular}{c}    \underline{0.289} \\ 3.845 \\ 0.617 \\ 1.13 \\ 51.221 \\ 26.846 \\ 100.497 \\  \end{tabular}
    \\
    \begin{tabular}{ll} Bikesharing \\ (Temperature) \end{tabular}
    & \renewcommand{\arraystretch}{0.8} \begin{tabular}{c} MDD \\ ACD \\ SD \\ KD \\ ED \\ DTW \\ SHAP-RE \end{tabular}
    & \renewcommand{\arraystretch}{0.8} \begin{tabular}{c}    0.37 \\ 1.426 \\ \underline{0.13} \\ 1.91 \\ \underline{17.172} \\ 12.107 \\ \underline{2.807} \\  \end{tabular}
    & \renewcommand{\arraystretch}{0.8} \begin{tabular}{c}    0.372 \\ 1.662 \\ 0.488 \\ 0.931 \\ 20.38 \\ 12.47 \\ 4.426 \\  \end{tabular}
    & \renewcommand{\arraystretch}{0.8} \begin{tabular}{c}    0.363 \\ \textbf{0.086} \\ 0.43 \\ \underline{0.536} \\ 19.349 \\ \underline{11.074} \\ 7.514 \\  \end{tabular}
    & \renewcommand{\arraystretch}{0.8} \begin{tabular}{c}    0.451 \\ 5.766 \\ 0.534 \\ 1.776 \\ \textbf{15.702} \\ \textbf{10.569} \\ \textbf{2.189} \\  \end{tabular}
    & \renewcommand{\arraystretch}{0.8} \begin{tabular}{c}    0.386 \\ 2.137 \\ 0.808 \\ 1.228 \\ 20.591 \\ 15.251 \\ 6.538 \\  \end{tabular}
    & \renewcommand{\arraystretch}{0.8} \begin{tabular}{c}    \textbf{0.291} \\ \underline{1.244} \\ \textbf{0.11} \\ \textbf{0.409} \\ 49.379 \\ 46.37 \\ 42.472 \\  \end{tabular}
    & \renewcommand{\arraystretch}{0.8} \begin{tabular}{c}    \underline{0.339} \\ 7.736 \\ 0.53 \\ 1.334 \\ 39.062 \\ 26.093 \\ 50.348 \\  \end{tabular}
    \\
    \begin{tabular}{ll} Traffic \\ (Junction 1) \end{tabular}
    & \renewcommand{\arraystretch}{0.8} \begin{tabular}{c} MDD \\ ACD \\ SD \\ KD \\ ED \\ DTW \\ SHAP-RE \end{tabular}
    & \renewcommand{\arraystretch}{0.8} \begin{tabular}{c}    0.3 \\ \underline{0.924} \\ 0.276 \\ 3.793 \\ 18.191 \\ 10.776 \\ 5.198 \\  \end{tabular}
    & \renewcommand{\arraystretch}{0.8} \begin{tabular}{c}    0.294 \\ \textbf{0.861} \\ \underline{0.018} \\ 2.151 \\ 18.465 \\ \underline{10.597} \\ 5.21 \\  \end{tabular}
    & \renewcommand{\arraystretch}{0.8} \begin{tabular}{c}    \underline{0.277} \\ 4.631 \\ 0.667 \\ 0.769 \\ 25.655 \\ 13.931 \\ 31.453 \\  \end{tabular}
    & \renewcommand{\arraystretch}{0.8} \begin{tabular}{c}    0.42 \\ 2.642 \\ \textbf{0.008} \\ 1.807 \\ \textbf{15.782} \\ 10.893 \\ \textbf{2.124} \\  \end{tabular}
    & \renewcommand{\arraystretch}{0.8} \begin{tabular}{c}    0.403 \\ 2.195 \\ 0.225 \\ \underline{0.317} \\ \underline{16.151} \\ \textbf{10.065} \\ \underline{4.393} \\  \end{tabular}
    & \renewcommand{\arraystretch}{0.8} \begin{tabular}{c}    \textbf{0.267} \\ 6.292 \\ 0.167 \\ \textbf{0.052} \\ 45.765 \\ 41.767 \\ 82.871 \\  \end{tabular}
    & \renewcommand{\arraystretch}{0.8} \begin{tabular}{c}    0.313 \\ 3.854 \\ 0.266 \\ 1.168 \\ 36.403 \\ 18.485 \\ 58.632 \\  \end{tabular}
    \\
    \begin{tabular}{ll} Traffic \\ (Junction 2) \end{tabular}
    & \renewcommand{\arraystretch}{0.8} \begin{tabular}{c} MDD \\ ACD \\ SD \\ KD \\ ED \\ DTW \\ SHAP-RE \end{tabular}
    & \renewcommand{\arraystretch}{0.8} \begin{tabular}{c}    0.367 \\ 2.893 \\ 0.272 \\ 1.212 \\ 17.937 \\ \textbf{10.246} \\ 5.546 \\  \end{tabular}
    & \renewcommand{\arraystretch}{0.8} \begin{tabular}{c}    0.368 \\ \textbf{2.265} \\ \underline{0.237} \\ \underline{0.645} \\ 18.765 \\ 10.425 \\ 6.777 \\  \end{tabular}
    & \renewcommand{\arraystretch}{0.8} \begin{tabular}{c}    0.352 \\ 2.889 \\ 0.526 \\ 1.135 \\ 21.622 \\ \underline{10.307} \\ 30.016 \\  \end{tabular}
    & \renewcommand{\arraystretch}{0.8} \begin{tabular}{c}    0.434 \\ 2.878 \\ \textbf{0.086} \\ 1.921 \\ \textbf{16.007} \\ 10.995 \\ \textbf{2.157} \\  \end{tabular}
    & \renewcommand{\arraystretch}{0.8} \begin{tabular}{c}    0.445 \\ \underline{2.516} \\ 0.992 \\ 2.667 \\ \underline{17.114} \\ 11.314 \\ \underline{3.426} \\  \end{tabular}
    & \renewcommand{\arraystretch}{0.8} \begin{tabular}{c}    \textbf{0.281} \\ 5.164 \\ 0.328 \\ \textbf{0.273} \\ 44.67 \\ 38.389 \\ 135.497 \\  \end{tabular}
    & \renewcommand{\arraystretch}{0.8} \begin{tabular}{c}    \underline{0.315} \\ 4.492 \\ 0.295 \\ 1.155 \\ 41.478 \\ 19.57 \\ 99.919 \\  \end{tabular}
    \\
    \begin{tabular}{ll} Traffic \\ (Junction 3) \end{tabular}
    & \renewcommand{\arraystretch}{0.8} \begin{tabular}{c} MDD \\ ACD \\ SD \\ KD \\ ED \\ DTW \\ SHAP-RE \end{tabular}
    & \renewcommand{\arraystretch}{0.8} \begin{tabular}{c}    0.326 \\ 2.788 \\ 0.898 \\ 1.086 \\ \underline{17.961} \\ 11.558 \\ \underline{6.848} \\  \end{tabular}
    & \renewcommand{\arraystretch}{0.8} \begin{tabular}{c}    0.326 \\ \underline{2.444} \\ 1.039 \\ 0.684 \\ 18.412 \\ 11.147 \\ 7.446 \\  \end{tabular}
    & \renewcommand{\arraystretch}{0.8} \begin{tabular}{c}    0.326 \\ 3.861 \\ \textbf{0.054} \\ \textbf{0.262} \\ 22.054 \\ \underline{10.892} \\ 46.675 \\  \end{tabular}
    & \renewcommand{\arraystretch}{0.8} \begin{tabular}{c}    0.41 \\ \textbf{2.123} \\ 0.906 \\ \underline{0.336} \\ \textbf{16.058} \\ 11.807 \\ \textbf{2.022} \\  \end{tabular}
    & \renewcommand{\arraystretch}{0.8} \begin{tabular}{c}    0.361 \\ 3.308 \\ \underline{0.23} \\ 0.776 \\ 18.829 \\ \textbf{10.677} \\ 12.436 \\  \end{tabular}
    & \renewcommand{\arraystretch}{0.8} \begin{tabular}{c}    \textbf{0.25} \\ 5.97 \\ 1.056 \\ 1.283 \\ 38.533 \\ 32.538 \\ 118.245 \\  \end{tabular}
    & \renewcommand{\arraystretch}{0.8} \begin{tabular}{c}    \underline{0.276} \\ 4.377 \\ 1.257 \\ 2.822 \\ 53.251 \\ 25.618 \\ 178.736 \\  \end{tabular}
    \\
    \bottomrule
\end{tabular}
\label{tab:univariate_per_data_transport}
\end{table}

\begin{table}[H]
\caption{\textbf{Univariate evaluation: similarity metrics reported for Nature datasets.}}
\vspace{0.1cm}
\centering
\tiny
\renewcommand{\arraystretch}{1.4}
\rowcolors{2}{gray!20}{white}
\begin{tabular}{lccccccccc}
    \toprule
    & 
    & \multicolumn{2}{c}{\textbf{SDForger Models}} & \multicolumn{2}{c}{\textbf{VAE Models}} & \multicolumn{2}{c}{\textbf{GAN Models}} & \multicolumn{1}{c}{\textbf{Others}} \\
    \cmidrule(lr){3-4} \cmidrule(lr){5-6} \cmidrule(lr){7-8} \cmidrule(lr){9-9}
    & 
    & \rotatebox{0}{ICA$_3$} & \rotatebox{0}{FPC$_3$}
    & \rotatebox{0}{TimeVAE} & \rotatebox{0}{TimeVQVAE}
    & \rotatebox{0}{RTSGAN} & \rotatebox{0}{SDEGAN} 
    & \rotatebox{0}{LS4} \\
    \midrule
    \begin{tabular}{ll} CCP \\ (CO2) \end{tabular}
    & \renewcommand{\arraystretch}{0.8} \begin{tabular}{c} MDD \\ ACD \\ SD \\ KD \\ ED \\ DTW \\ SHAP-RE \end{tabular}
    & \renewcommand{\arraystretch}{0.8} \begin{tabular}{c}    \underline{0.173} \\ \underline{1.391} \\ \textbf{0.472} \\ \textbf{1.644} \\ 18.442 \\ 15.245 \\ \textbf{1.123} \\  \end{tabular}
    & \renewcommand{\arraystretch}{0.8} \begin{tabular}{c}    0.186 \\ \textbf{1.119} \\ 2.21 \\ 3.443 \\ 18.729 \\ 14.344 \\ 1.684 \\  \end{tabular}
    & \renewcommand{\arraystretch}{0.8} \begin{tabular}{c}    \textbf{0.166} \\ 2.203 \\ 2.481 \\ 4.155 \\ 26.232 \\ 18.308 \\ 13.719 \\  \end{tabular}
    & \renewcommand{\arraystretch}{0.8} \begin{tabular}{c}    0.3 \\ 5.536 \\ 1.984 \\ \underline{3.31} \\ \textbf{14.19} \\ \textbf{10.176} \\ 2.486 \\  \end{tabular}
    & \renewcommand{\arraystretch}{0.8} \begin{tabular}{c}    0.179 \\ 2.598 \\ \underline{1.12} \\ 3.72 \\ \underline{14.734} \\ \underline{10.831} \\ \underline{1.329} \\  \end{tabular}
    & \renewcommand{\arraystretch}{0.8} \begin{tabular}{c}    0.232 \\ 3.06 \\ 1.809 \\ 5.09 \\ 72.136 \\ 71.196 \\ 52.729 \\  \end{tabular}
    & \renewcommand{\arraystretch}{0.8} \begin{tabular}{c}    0.236 \\ 6.866 \\ 1.86 \\ 6.164 \\ 36.803 \\ 28.544 \\ 25.99 \\  \end{tabular} 
    \\
    \begin{tabular}{ll} CCP \\ (NH3) \end{tabular}
    & \renewcommand{\arraystretch}{0.8} \begin{tabular}{c} MDD \\ ACD \\ SD \\ KD \\ ED \\ DTW \\ SHAP-RE \end{tabular}
    & \renewcommand{\arraystretch}{0.8} \begin{tabular}{c}    0.452 \\ \textbf{0.312} \\ 0.416 \\ 1.267 \\ \underline{16.965} \\ 15.495 \\ \textbf{0.423} \\  \end{tabular}
    & \renewcommand{\arraystretch}{0.8} \begin{tabular}{c}    0.477 \\ 2.223 \\ 0.766 \\ 0.554 \\ 19.28 \\ 17.18 \\ \underline{1.064} \\  \end{tabular}
    & \renewcommand{\arraystretch}{0.8} \begin{tabular}{c}    0.409 \\ \underline{0.55} \\ \underline{0.272} \\ \underline{0.212} \\ 17.18 \\ \underline{14.483} \\ 1.13 \\  \end{tabular}
    & \renewcommand{\arraystretch}{0.8} \begin{tabular}{c}    0.67 \\ 6.779 \\ 0.441 \\ 0.292 \\ \textbf{15.465} \\ \textbf{12.381} \\ 2.418 \\  \end{tabular}
    & \renewcommand{\arraystretch}{0.8} \begin{tabular}{c}    \underline{0.392} \\ 0.56 \\ \textbf{0.23} \\ 0.671 \\ 22.525 \\ 19.973 \\ 2.904 \\  \end{tabular}
    & \renewcommand{\arraystretch}{0.8} \begin{tabular}{c}    \textbf{0.364} \\ 0.822 \\ 0.346 \\ \textbf{0.011} \\ 31.041 \\ 30.589 \\ 2.805 \\  \end{tabular}
    & \renewcommand{\arraystretch}{0.8} \begin{tabular}{c}    0.455 \\ 9.168 \\ 0.753 \\ 2.743 \\ 35.404 \\ 29.251 \\ 9.832 \\  \end{tabular}
    \\
    \begin{tabular}{ll} CCP \\ (C4H11NO) \end{tabular}
    & \renewcommand{\arraystretch}{0.8} \begin{tabular}{c} MDD \\ ACD \\ SD \\ KD \\ ED \\ DTW \\ SHAP-RE \end{tabular}
    & \renewcommand{\arraystretch}{0.8} \begin{tabular}{c}    0.197 \\ 0.662 \\ 1.747 \\ \underline{2.054} \\ 15.92 \\ 12.168 \\ \textbf{1.255} \\  \end{tabular}
    & \renewcommand{\arraystretch}{0.8} \begin{tabular}{c}    0.191 \\ 0.822 \\ 2.232 \\ 3.638 \\ 16.128 \\ 11.741 \\ 2.426 \\  \end{tabular}
    & \renewcommand{\arraystretch}{0.8} \begin{tabular}{c}    \textbf{0.158} \\ \textbf{0.152} \\ \textbf{0.007} \\ 3.105 \\ 15.448 \\ 10.881 \\ 2.801 \\  \end{tabular}
    & \renewcommand{\arraystretch}{0.8} \begin{tabular}{c}    0.266 \\ 6.672 \\ 2.294 \\ 4.998 \\ \textbf{11.295} \\ \textbf{8.252} \\ 2.363 \\  \end{tabular}
    & \renewcommand{\arraystretch}{0.8} \begin{tabular}{c}    \underline{0.189} \\ 0.371 \\ \underline{0.626} \\ \textbf{0.251} \\ \underline{14.54} \\ \underline{10.213} \\ \underline{2.042} \\  \end{tabular}
    & \renewcommand{\arraystretch}{0.8} \begin{tabular}{c}    \underline{0.189} \\ \underline{0.231} \\ 2.729 \\ 6.454 \\ 38.679 \\ 37.706 \\ 14.476 \\  \end{tabular}
    & \renewcommand{\arraystretch}{0.8} \begin{tabular}{c}    0.229 \\ 8.524 \\ 2.511 \\ 8.179 \\ 49.348 \\ 40.586 \\ 47.176 \\  \end{tabular}
    \\    
    \begin{tabular}{ll} CCP \\ (C4H10N2) \end{tabular}
    & \renewcommand{\arraystretch}{0.8} \begin{tabular}{c} MDD \\ ACD \\ SD \\ KD \\ ED \\ DTW \\ SHAP-RE \end{tabular}
    & \renewcommand{\arraystretch}{0.8} \begin{tabular}{c}    \underline{0.178} \\ 1.514 \\ 1.675 \\ \textbf{0.157} \\ \underline{15.698} \\ \underline{11.646} \\ \underline{2.299} \\  \end{tabular}
    & \renewcommand{\arraystretch}{0.8} \begin{tabular}{c}    0.18 \\ \underline{1.273} \\ 1.638 \\ 2.451 \\ 17.378 \\ 12.004 \\ 4.547 \\  \end{tabular}
    & \renewcommand{\arraystretch}{0.8} \begin{tabular}{c}    \textbf{0.162} \\ \textbf{0.535} \\ \textbf{0.512} \\ \underline{0.824} \\ 19.159 \\ 12.098 \\ 6.142 \\  \end{tabular}
    & \renewcommand{\arraystretch}{0.8} \begin{tabular}{c}    0.249 \\ 4.522 \\ 1.456 \\ 3.091 \\ \textbf{13.879} \\ \textbf{10.002} \\ \textbf{2.228} \\  \end{tabular}
    & \renewcommand{\arraystretch}{0.8} \begin{tabular}{c}    0.221 \\ 2.445 \\ \underline{1.398} \\ 4.551 \\ 21.6 \\ 14.647 \\ 5.582 \\  \end{tabular}
    & \renewcommand{\arraystretch}{0.8} \begin{tabular}{c}    0.19 \\ 3.142 \\ 1.831 \\ 4.592 \\ 46.177 \\ 44.366 \\ 31.911 \\  \end{tabular}
    & \renewcommand{\arraystretch}{0.8} \begin{tabular}{c}    0.249 \\ 5.538 \\ 1.659 \\ 6.204 \\ 45.823 \\ 33.614 \\ 59.928 \\  \end{tabular}
    \\
    \bottomrule
\end{tabular}
\label{tab:univariate_per_data_nature}
\end{table}

\begin{table}[H]
\caption{\textbf{Univariate evaluation: similarity metrics reported for Finance datasets.}}
\vspace{0.1cm}
\centering
\tiny
\renewcommand{\arraystretch}{1.4}
\rowcolors{2}{gray!20}{white}
\begin{tabular}{lccccccccc}
    \toprule
    & 
    & \multicolumn{2}{c}{\textbf{SDForger Models}} & \multicolumn{2}{c}{\textbf{VAE Models}} & \multicolumn{2}{c}{\textbf{GAN Models}} & \multicolumn{1}{c}{\textbf{Others}} \\
    \cmidrule(lr){3-4} \cmidrule(lr){5-6} \cmidrule(lr){7-8} \cmidrule(lr){9-9}
    & 
    & \rotatebox{0}{ICA$_3$} & \rotatebox{0}{FPC$_3$}
    & \rotatebox{0}{TimeVAE} & \rotatebox{0}{TimeVQVAE}
    & \rotatebox{0}{RTSGAN} & \rotatebox{0}{SDEGAN} 
    & \rotatebox{0}{LS4} \\
    \midrule
    \begin{tabular}{ll} Exchange Rate \\ (Currency 1) \end{tabular}
    & \renewcommand{\arraystretch}{0.8} \begin{tabular}{c} MDD \\ ACD \\ SD \\ KD \\ ED \\ DTW \\ SHAP-RE \end{tabular}
    & \renewcommand{\arraystretch}{0.8} \begin{tabular}{c}    0.27 \\ \underline{0.851} \\ 0.409 \\ 0.32 \\ \underline{18.913} \\ \underline{15.672} \\ \textbf{1.759} \\  \end{tabular}
    & \renewcommand{\arraystretch}{0.8} \begin{tabular}{c}    \underline{0.269} \\ 1.727 \\ 0.62 \\ 1.173 \\ 19.909 \\ 15.91 \\ \underline{2.228} \\  \end{tabular}
    & \renewcommand{\arraystretch}{0.8} \begin{tabular}{c}    \textbf{0.267} \\ \textbf{0.211} \\ \textbf{0.33} \\ \underline{0.058} \\ 20.652 \\ 16.319 \\ 2.52 \\  \end{tabular}
    & \renewcommand{\arraystretch}{0.8} \begin{tabular}{c}    0.415 \\ 7.212 \\ 0.614 \\ 1.77 \\ \textbf{15.335} \\ \textbf{11.256} \\ 2.449 \\  \end{tabular}
    & \renewcommand{\arraystretch}{0.8} \begin{tabular}{c}    0.358 \\ 1.062 \\ 1.031 \\ 0.854 \\ 26.589 \\ 22.394 \\ 5.691 \\  \end{tabular}
    & \renewcommand{\arraystretch}{0.8} \begin{tabular}{c}    0.278 \\ 0.932 \\ 0.476 \\ \textbf{0.034} \\ 47.853 \\ 46.856 \\ 17.248 \\  \end{tabular}
    & \renewcommand{\arraystretch}{0.8} \begin{tabular}{c}    0.308 \\ 8.803 \\ \underline{0.332} \\ 1.275 \\ 35.288 \\ 26.695 \\ 22.698 \\  \end{tabular}
    
    \\
    \begin{tabular}{ll} Exchange Rate \\ (Currency 2) \end{tabular}
    & \renewcommand{\arraystretch}{0.8} \begin{tabular}{c} MDD \\ ACD \\ SD \\ KD \\ ED \\ DTW \\ SHAP-RE \end{tabular}
    & \renewcommand{\arraystretch}{0.8} \begin{tabular}{c}    0.316 \\ \underline{0.426} \\ 1.285 \\ 2.003 \\ \underline{17.727} \\ 15.12 \\ \textbf{0.967} \\  \end{tabular}
    & \renewcommand{\arraystretch}{0.8} \begin{tabular}{c}    0.303 \\ 1.424 \\ 0.914 \\ 1.249 \\ 19.579 \\ 16.076 \\ \underline{1.186} \\  \end{tabular}
    & \renewcommand{\arraystretch}{0.8} \begin{tabular}{c}    \underline{0.286} \\ \textbf{0.305} \\ \textbf{0.289} \\ 1.507 \\ 18.253 \\ \underline{14.15} \\ 1.744 \\  \end{tabular}
    & \renewcommand{\arraystretch}{0.8} \begin{tabular}{c}    0.459 \\ 7.14 \\ 0.863 \\ 1.576 \\ \textbf{14.43} \\ \textbf{10.689} \\ 2.442 \\  \end{tabular}
    & \renewcommand{\arraystretch}{0.8} \begin{tabular}{c}    0.343 \\ 0.983 \\ \underline{0.424} \\ \underline{0.443} \\ 18.272 \\ 14.488 \\ 2.217 \\  \end{tabular}
    & \renewcommand{\arraystretch}{0.8} \begin{tabular}{c}    \textbf{0.253} \\ 0.976 \\ 1.034 \\ \textbf{0.208} \\ 32.215 \\ 31.633 \\ 6.681 \\  \end{tabular}
    & \renewcommand{\arraystretch}{0.8} \begin{tabular}{c}    0.291 \\ 8.597 \\ 1.083 \\ 1.636 \\ 45.617 \\ 38.042 \\ 28.819 \\  \end{tabular}
    
    \\
    \begin{tabular}{ll} Exchange Rate \\ (Currency 3) \end{tabular}
    & \renewcommand{\arraystretch}{0.8} \begin{tabular}{c} MDD \\ ACD \\ SD \\ KD \\ ED \\ DTW \\ SHAP-RE \end{tabular}
    & \renewcommand{\arraystretch}{0.8} \begin{tabular}{c}    0.369 \\ \underline{0.113} \\ \underline{0.22} \\ 0.64 \\ 19.578 \\ 19.092 \\ \textbf{0.485} \\  \end{tabular}
    & \renewcommand{\arraystretch}{0.8} \begin{tabular}{c}    0.428 \\ 2.852 \\ 0.751 \\ 3.039 \\ 21.042 \\ 19.25 \\ 0.956 \\  \end{tabular}
    & \renewcommand{\arraystretch}{0.8} \begin{tabular}{c}    0.36 \\ 0.157 \\ \textbf{0.122} \\ \underline{0.205} \\ \underline{19.31} \\ \underline{18.532} \\ \underline{0.808} \\  \end{tabular}
    & \renewcommand{\arraystretch}{0.8} \begin{tabular}{c}    0.614 \\ 8.035 \\ 0.453 \\ 2.755 \\ \textbf{14.765} \\ \textbf{12.47} \\ 2.44 \\  \end{tabular}
    & \renewcommand{\arraystretch}{0.8} \begin{tabular}{c}    0.47 \\ 0.639 \\ 1.154 \\ \textbf{0.046} \\ 21.985 \\ 20.528 \\ 1.168 \\  \end{tabular}
    & \renewcommand{\arraystretch}{0.8} \begin{tabular}{c}    \textbf{0.314} \\ \textbf{0.106} \\ 0.633 \\ 0.992 \\ 28.394 \\ 28.143 \\ 2.133 \\  \end{tabular}
    & \renewcommand{\arraystretch}{0.8} \begin{tabular}{c}    \underline{0.348} \\ 9.934 \\ 0.71 \\ 0.369 \\ 38.289 \\ 34.966 \\ 15.265 \\  \end{tabular}
    
    \\
    \begin{tabular}{ll} Exchange Rate \\ (Currency 4) \end{tabular}
    & \renewcommand{\arraystretch}{0.8} \begin{tabular}{c} MDD \\ ACD \\ SD \\ KD \\ ED \\ DTW \\ SHAP-RE \end{tabular}
    & \renewcommand{\arraystretch}{0.8} \begin{tabular}{c}    0.342 \\ \textbf{0.34} \\ 0.647 \\ \underline{0.484} \\ 20.09 \\ 16.021 \\ \textbf{1.18} \\  \end{tabular}
    & \renewcommand{\arraystretch}{0.8} \begin{tabular}{c}    0.341 \\ 1.427 \\ \underline{0.252} \\ 1.206 \\ 20.287 \\ 16.187 \\ \underline{1.383} \\  \end{tabular}
    & \renewcommand{\arraystretch}{0.8} \begin{tabular}{c}    0.341 \\ \underline{0.848} \\ \textbf{0.136} \\ 0.69 \\ \underline{19.622} \\ \underline{13.507} \\ 1.85 \\  \end{tabular}
    & \renewcommand{\arraystretch}{0.8} \begin{tabular}{c}    0.543 \\ 7.171 \\ 0.349 \\ 2.483 \\ \textbf{15.38} \\ \textbf{10.985} \\ 2.34 \\  \end{tabular}
    & \renewcommand{\arraystretch}{0.8} \begin{tabular}{c}    0.349 \\ 1.132 \\ 0.49 \\ \textbf{0.394} \\ 20.51 \\ 14.745 \\ 1.858 \\  \end{tabular}
    & \renewcommand{\arraystretch}{0.8} \begin{tabular}{c}    \textbf{0.279} \\ 1.023 \\ 0.502 \\ 0.666 \\ 33.675 \\ 32.558 \\ 7.303 \\  \end{tabular}
    & \renewcommand{\arraystretch}{0.8} \begin{tabular}{c}    \underline{0.317} \\ 8.913 \\ 0.686 \\ 0.645 \\ 42.123 \\ 33.075 \\ 23.273 \\  \end{tabular}
    
    \\
    \begin{tabular}{ll} Exchange Rate \\ (Currency 6) \end{tabular}
    & \renewcommand{\arraystretch}{0.8} \begin{tabular}{c} MDD \\ ACD \\ SD \\ KD \\ ED \\ DTW \\ SHAP-RE \end{tabular}
    & \renewcommand{\arraystretch}{0.8} \begin{tabular}{c}    0.273 \\ \underline{0.204} \\ 0.285 \\ \underline{0.534} \\ 19.809 \\ 18.469 \\ \textbf{0.569} \\  \end{tabular}
    & \renewcommand{\arraystretch}{0.8} \begin{tabular}{c}    0.28 \\ 1.893 \\ 0.755 \\ 3.95 \\ 19.763 \\ 17.954 \\ \underline{0.82} \\  \end{tabular}
    & \renewcommand{\arraystretch}{0.8} \begin{tabular}{c}    \underline{0.251} \\ \textbf{0.179} \\ \underline{0.245} \\ 0.88 \\ \underline{16.956} \\ \underline{14.799} \\ 1.036 \\  \end{tabular}
    & \renewcommand{\arraystretch}{0.8} \begin{tabular}{c}    0.427 \\ 7.773 \\ 0.342 \\ 1.701 \\ \textbf{14.579} \\ \textbf{11.822} \\ 2.386 \\  \end{tabular}
    & \renewcommand{\arraystretch}{0.8} \begin{tabular}{c}    0.338 \\ 0.578 \\ \textbf{0.049} \\ 1.127 \\ 21.034 \\ 18.02 \\ 2.427 \\  \end{tabular}
    & \renewcommand{\arraystretch}{0.8} \begin{tabular}{c}    \textbf{0.237} \\ 0.414 \\ 0.481 \\ \textbf{0.152} \\ 39.025 \\ 38.671 \\ 5.158 \\  \end{tabular}
    & \renewcommand{\arraystretch}{0.8} \begin{tabular}{c}    0.286 \\ 8.341 \\ 0.464 \\ 1.401 \\ 34.643 \\ 30.317 \\ 28.882 \\  \end{tabular}
    
    \\
    \begin{tabular}{ll} Exchange Rate \\ (Currency 7) \end{tabular}
    & \renewcommand{\arraystretch}{0.8} \begin{tabular}{c} MDD \\ ACD \\ SD \\ KD \\ ED \\ DTW \\ SHAP-RE \end{tabular}
    & \renewcommand{\arraystretch}{0.8} \begin{tabular}{c}    \underline{0.318} \\ \underline{0.17} \\ 0.249 \\ 0.743 \\ 19.565 \\ 18.484 \\ 0.945 \\  \end{tabular}
    & \renewcommand{\arraystretch}{0.8} \begin{tabular}{c}    0.353 \\ 2.423 \\ \textbf{0.169} \\ 3.389 \\ 23.189 \\ 20.773 \\ \textbf{0.761} \\  \end{tabular}
    & \renewcommand{\arraystretch}{0.8} \begin{tabular}{c}    0.322 \\ 0.179 \\ \underline{0.212} \\ 0.574 \\ \underline{15.034} \\ \underline{13.105} \\ 0.981 \\  \end{tabular}
    & \renewcommand{\arraystretch}{0.8} \begin{tabular}{c}    0.493 \\ 7.967 \\ 0.22 \\ 2.616 \\ \textbf{14.768} \\ \textbf{12.875} \\ 2.496 \\  \end{tabular}
    & \renewcommand{\arraystretch}{0.8} \begin{tabular}{c}    0.374 \\ \textbf{0.121} \\ 0.61 \\ \textbf{0.379} \\ 21.199 \\ 19.659 \\ \underline{0.918} \\  \end{tabular}
    & \renewcommand{\arraystretch}{0.8} \begin{tabular}{c}    \textbf{0.292} \\ 0.22 \\ 0.344 \\ 0.759 \\ 36.463 \\ 36.214 \\ 2.649 \\  \end{tabular}
    & \renewcommand{\arraystretch}{0.8} \begin{tabular}{c}    0.334 \\ 9.778 \\ 0.599 \\ \underline{0.443} \\ 34.437 \\ 30.555 \\ 9.963 \\  \end{tabular}
    
    \\
    \begin{tabular}{ll} Exchange Rate \\ (Currency 8) \end{tabular} 
    & \renewcommand{\arraystretch}{0.8} \begin{tabular}{c} MDD \\ ACD \\ SD \\ KD \\ ED \\ DTW \\ SHAP-RE \end{tabular}
    & \renewcommand{\arraystretch}{0.8} \begin{tabular}{c}    0.401 \\ \textbf{0.06} \\ 0.091 \\ 1.009 \\ \underline{16.88} \\ 16.254 \\ \textbf{0.232} \\  \end{tabular}
    & \renewcommand{\arraystretch}{0.8} \begin{tabular}{c}    0.378 \\ 2.259 \\ \textbf{0.04} \\ 2.867 \\ 21.571 \\ 20.245 \\ 1.044 \\  \end{tabular}
    & \renewcommand{\arraystretch}{0.8} \begin{tabular}{c}    0.358 \\ 0.119 \\ 0.595 \\ 0.716 \\ 16.902 \\ \underline{15.651} \\ \underline{0.638} \\  \end{tabular}
    & \renewcommand{\arraystretch}{0.8} \begin{tabular}{c}    0.564 \\ 8.052 \\ 0.173 \\ 2.751 \\ \textbf{14.882} \\ \textbf{12.884} \\ 2.43 \\  \end{tabular}
    & \renewcommand{\arraystretch}{0.8} \begin{tabular}{c}    0.467 \\ 0.197 \\ 0.253 \\ \textbf{0.202} \\ 21.432 \\ 20.061 \\ 0.679 \\  \end{tabular}
    & \renewcommand{\arraystretch}{0.8} \begin{tabular}{c}    \textbf{0.339} \\ \underline{0.077} \\ \underline{0.041} \\ 0.891 \\ 37.487 \\ 37.232 \\ 3.228 \\  \end{tabular}
    & \renewcommand{\arraystretch}{0.8} \begin{tabular}{c}    \underline{0.353} \\ 10.102 \\ 0.333 \\ \underline{0.205} \\ 42.048 \\ 39.402 \\ 15.895 \\  \end{tabular}
    \\
    \bottomrule
\end{tabular}
\label{tab:univariate_per_data_finance}
\end{table}


\end{document}